\documentclass[preprint,5p, 10pt, times]{elsarticle}

\usepackage{color, colortbl}

\usepackage{multirow}
\usepackage{makecell}
\usepackage{array}
\usepackage{pifont}
\usepackage{ifsym}
\usepackage{tabu}

\usepackage{upquote}

\usepackage{mdframed}
\usepackage[dvipsnames, table]{xcolor}
\usepackage{amssymb}
\usepackage{color}
\usepackage{subcaption}
\definecolor{Green}{RGB}{0,102,102}
\usepackage{tikz}

\journal{~}
\usepackage{titlesec}

\titleformat{\section}
{\color{black}\normalfont\large\bf\sffamily}
{\color{black}\thesection}{1em}{}
\usepackage[colorlinks,citecolor=blue,urlcolor=blue,bookmarks=false,hypertexnames=true,breaklinks=true]{hyperref}
\titleformat{\subsection}
{\color{black}\normalfont\normalsize\bf\sffamily}
{\color{black}\thesubsection}{1em}{}{}

\usepackage[superscript,biblabel, compress]{cite}
\usepackage{titletoc}

\usepackage[labeled,resetlabels]{multibib}
\newcites{a}{ }

\usepackage{tabu}

\usepackage{soul}

\usepackage{booktabs}

\usepackage{graphicx}
\usepackage{subcaption}

\usepackage{xr}
\usepackage{multirow}

\usepackage{stfloats}

\usepackage{pdfpages}

\begin{document}
\sloppy

\begin{frontmatter}

\title{\LARGE \bf Accuracy and Consistency of LLMs in the Registered Dietitian Exam: \\The Impact of Prompt Engineering and Knowledge Retrieval}
%\title{\LARGE \bf (Tentative Title) Evaluation of Large Language Models on the Registered Dietitian Exam: Can Reasoning or Knowledge Retrieval Enhance Performance?}
% A title that is up to 15 words in length and free of punctuation, idioms, and puns; the full author list; author affiliation information, including institution, city, and country; and the corresponding author(s) email address

\author{\textbf{\fontsize{11pt}{13.3pt}\selectfont{Iman Azimi\textsuperscript{1,*}, Mohan Qi\textsuperscript{1}, Li Wang\textsuperscript{2},  Amir M. Rahmani\textsuperscript{3}, and Youlin Li\textsuperscript{1}}}~\\\small\normalfont 
~\\\textsuperscript{1}{Department of Engineering, iHealth Labs}
~\\\textsuperscript{2}{Department of Clinical Research, iHealth Labs}
~\\\textsuperscript{3}{School of Nursing and Department of Computer Science, University of California, Irvine}
~\\\textsuperscript{*}{Corresponding author, iman.azimi@ihealthlabs.com}}

\begin{abstract}
% No subheadings permitted and up to 150 words in length
\centering\begin{minipage}{\dimexpr\paperwidth-3cm}

Large language models (LLMs) are fundamentally transforming human-facing applications in the health and well-being domains: boosting patient engagement, accelerating clinical decision-making, and facilitating medical education. Although state-of-the-art LLMs have shown superior performance in several conversational applications, evaluations within nutrition and diet applications are still insufficient. In this paper, we propose to employ the Registered Dietitian (RD) exam to conduct a standard and comprehensive evaluation of state-of-the-art LLMs, GPT-4o, Claude 3.5 Sonnet, and Gemini 1.5 Pro, assessing both accuracy and consistency in nutrition queries. Our evaluation includes 1050 RD exam questions encompassing several nutrition topics and proficiency levels. In addition, for the first time, we examine the impact of Zero-Shot (ZS), Chain of Thought (CoT), Chain of Thought with Self Consistency (CoT-SC), and Retrieval Augmented Prompting (RAP) on both accuracy and consistency of the responses. Our findings revealed that while these LLMs obtained acceptable overall performance, their results varied considerably with different prompts and question domains. GPT-4o with CoT-SC prompting outperformed the other approaches, whereas Gemini 1.5 Pro with ZS recorded the highest consistency. For GPT-4o and Claude 3.5, CoT improved the accuracy, and CoT-SC improved both accuracy and consistency. RAP was particularly effective for GPT-4o to answer Expert level questions. Consequently, choosing the appropriate LLM and prompting technique, tailored to the proficiency level and specific domain, can mitigate errors and potential risks in diet and nutrition chatbots.

\end{minipage}

\end{abstract}

\end{frontmatter}

\section*{Introduction}
\label{Sec:introduction}
% No subheadings permitted

There is growing interest in leveraging conversational models, commonly known as chatbots, in healthcare, particularly in the areas of diet and nutrition \cite{singh2023systematic,webster2023six,ma2024large}. The rise of large language models (LLMs) is significantly transforming human-machine interactions in this context, creating new opportunities for nutrition management applications and lifestyle enhancement that involve natural language understanding and generation \cite{clusmann2023future,mesko2023impact,bond2023artificial}. These chatbots can serve as assistants to health providers (e.g., dietitian or nurses) or as ubiquitous companions for patients, providing preventive care, personalized meal planning, and chronic disease management \cite{dao2024llm}. 

Since the release of ChatGPT \cite{openai2022chatgpt} in November 2022, numerous nutrition management studies have developed or employed LLM-based chatbots to target different health conditions, such as type 2 diabetes, obesity, liver diseases, kidney diseases, and cardiovascular diseases, to mention a few \cite{singh2023systematic,liu2023exploring,pugliese2024accuracy,kim2024qualitative,dao2024llm,abbasian2024knowledge,haman2024ai,qarajeh2023ai,tsai2023generating,zhou2024foodsky}. These studies highlight the potential of chatbots interventions to enhance diet and promote lifestyle behavior changes.

Due to the life-critical nature of these applications, they must provide high quality attributes, such as accuracy, consistency, safety, and fairness, before being deployed in real-world settings for end-users \cite{thirunavukarasu2023large,abbasian2024foundation,liang2022holistic}. Recent studies have evaluated the LLM-based chatbots within nutritional and dietary contexts. For example, Sun \textit{et al.} \cite{sun2023ai} and Barlas \textit{et al.} \cite{barlas2024credibility} assessed the performance of ChatGPT in providing nutritional management support for diabetic patients. Other investigations focused on chatbots' reliability in delivering accurate calorie and macronutrient information \cite{hoang2023consistency,yang2024chatdiet}. For non-communicable diseases, the accuracy of dietary advice generated by ChatGPT's were assessed \cite{ponzo2024chatgpt,pugliese2024accuracy}. Other studies also examined ChatGPT's ability to address common nutrition-related inquiries, highlighting its strength and weakness in offering personalized and accurate nutritional information \cite{kirk2023comparison,szymanski2024integrating}. However, the existing evaluation studies on nutrition-related chatbots face three major challenges. 

First, prior research on the LLMs application in nutrition has relied solely on ad-hoc or subjective evaluations. In these studies, domain experts designed a set of questions focused on specific diseases or nutrition topics. Subsequently, human evaluators were instructed to grade the responses in terms of accuracy, comprehensiveness, or attractiveness \cite{niszczota2023credibility,barlas2024credibility,sun2023ai}. Human-in-the-loop evaluation is widely recognized as a popular and well-established strategy for assessing chatbots in the literature \cite{abbasian2024foundation,liang2022holistic}. However, these evaluations are not comprehensive regarding nutrition problems and are prone to human errors or biases, as they depend on the opinion of an individual expert, especially when no standard guidelines are followed in the evaluation process. Additionally, they are time-consuming and costly. This limitation can be observed in the current nutrition chatbots evaluation, as their assessments are restricted to a few hundred interactions (i.e., prompts) at most.

Second, most of the nutrition and diet studies have focused only on ChatGPT-3.5 or ChatGPT-4. The landscape of LLMs is rapidly evolving. New models and techniques are being released frequently, within weeks or months \cite{minaee2024large}. This rapid advancement requires the evaluation of a wide range of models to ensure the best possible solutions for diet and nutrition management applications. In addition, existing research on nutrition evaluation has ignored the impact of prompt engineering techniques. They have been limited to zero-shot prompting methods with either no instructions or fixed instructions. Prompt engineering is an important technique for enhancing the capabilities, adaptability, and applicability of LLMs \cite{sahoo2024systematic,chen2023unleashing,wang2023prompt,maharjan2024openmedlm}. 
%It aims to create prompts that guide LLMs to produce desired outputs without changing the models' parameters \cite{wang2024prompt,maharjan2024openmedlm,wang2023prompt}. 
%It is essential to investigate how prompt engineering can affect LLM responses. Understanding their impact could lead to more effective use of LLMs across nutrition and diet applications, ultimately improving their performance and reliability.

Third, previous work merely focused on the overall accuracy of LLMs responses. Their findings indicated that the models were generally accurate, but they still had errors \cite{pugliese2024accuracy, niszczota2023credibility,barlas2024credibility,ponzo2024chatgpt}. These studies did not examine the errors, along with the strategies to enhance the LLMs' responses. Wang \textit{et al.} \cite{wang2024prompt} highlights this issue in the context of clinical medicine. Moreover, the non-deterministic behavior of LLMs was ignored \cite{ouyang2023llm}. Within the healthcare and medical sectors, there is a strong demand for deterministic outcomes, ensuring that identical inputs generate identical outputs. The consistency and reliability of LLMs in answering nutrition-related questions must be evaluated to determine if their performance varies with identical or different prompts. In the nutrition context, to the best of our knowledge, only one study \cite{hoang2023consistency} has explored the consistency of ChatGPT-3.5 and ChatGPT-4 responses, using a zero-shot prompt for 222 food items across five repeated measurements.

\begin{figure*}[bp]
    \centering
    \includegraphics[width=0.75\textwidth]{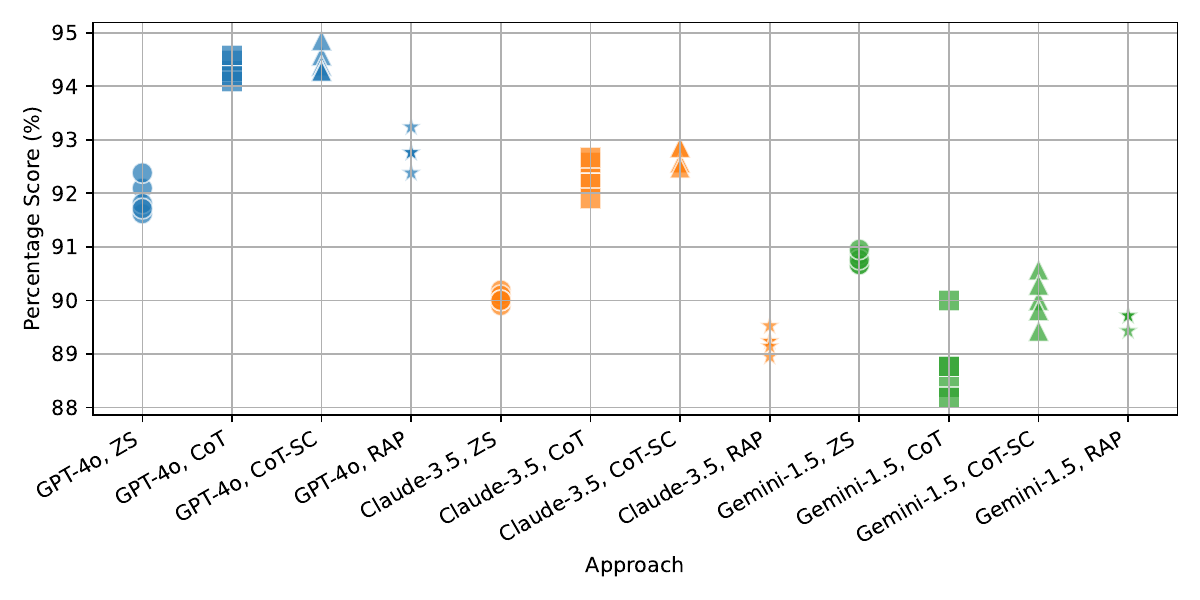}
    \caption{Percentage Scores of the approaches on the RD exam. GPT-4o, Claude 3.5 Sonnet, and Gemini 1.5 Pro are indicated with blue, orange, and green markers, respectively. The Zero Shot (ZS), Chain of Thought (CoT), Chain of Thought with Self Consistency (CoT-SC), and Retrieval Augmented Prompting (RAP) techniques are indicated with circle, square, triangle, and star markers, respectively.}
    \label{fig:accuracy_percentage}
\end{figure*}

In this paper, we thoroughly evaluate the accuracy and consistency of GPT-4o \cite{openai2024gpt4}, Claude 3.5 Sonnet \cite{anthropic2024claude}, and Gemini 1.5 Pro \cite{google2024gemini} in addressing nutrition-related inquiries. To achieve this, we leverage the Registered Dietitian (RD) exam \cite{cdr2024rdexam} for the first time, as a standard certification examination that serves to assess whether dietitians meet the qualifications required to practice in the dietetics and nutrition field. Our evaluation includes 1050 multiple-choice questions with different proficiency levels, covering four nutrition domains: i.e., principles of dietetics, nutrition care, food service systems, and food and nutrition management. To investigate the impact of prompts, the questions are presented to the LLMs using four different prompting techniques: 1) Zero Shot prompting (ZS), 2) Chain of Thought (CoT), 3) Chain of Thought with Self Consistency (CoT-SC), and 4) Retrieval Augmented Prompting (RAP) enabled by external nutrition knowledge. We then compare the responses with the ground truth answers, enabling an objective assessment of the model's performance. To examine the consistency of the responses, we perform repeated measurements by asking each model the same set of questions multiple times using each prompting technique. The responses for each technique and model are compared within and across groups.

\begin{table*}[!t]
\caption{The percentage scores (mean and standard deviation) of the LLMs' responses on the RD exam questions.}
\label{table:accuracy_percentage}
\centering
\small
\begin{tabular}{@{}>{\raggedright}m{0.18\textwidth}>{\raggedright}p{0.30\textwidth}>
{\centering\arraybackslash}m{0.13\textwidth}>{\centering\arraybackslash}m{0.13\textwidth}>{\centering\arraybackslash}m{0.13\textwidth}@{}}
\toprule
Benchmark & Prompt & {GPT-4o} & {Claude 3.5~S.} & {Gemini 1.5~P.} \\ 
\toprule
\multirow{4}{*}{RD Exam} & Zero Shot & \textbf{91.92\% (0.28)} & 90.04\% (0.10)& 90.78\% (0.11)\\
& Chain of Thought & \textbf{94.32\% (0.18)} & 92.32\% (0.27) & 88.82\% (0.63)\\
& Chain of Thought w. Self Consistency & \textbf{94.48\% (0.22)} & 92.67\% (0.16) & 90.02\% (0.39)\\
& Retrieval Augmented Prompting & \textbf{92.78\% (0.27)} & 89.22\% (0.18) & 89.66\% (0.11)\\
\bottomrule
\end{tabular}
\end{table*}

\begin{table*}[!t]
\caption{The performance of the LLMs on the MMLU \cite{hendrycks2020measuring}, GPQA \cite{rein2023gpqa}, and DROP \cite{dua2019drop} benchmarks, collected from \cite{anthropic2024claude,openai2024gpt4,reid2024gemini}.}
\label{table:accuracy_percentage_other_benchmarks}
\centering
\small
\begin{tabular}{@{}>{\raggedright}m{0.18\textwidth}>{\raggedright}p{0.30\textwidth}>
{\centering\arraybackslash}m{0.13\textwidth}>{\centering\arraybackslash}m{0.13\textwidth}>{\centering\arraybackslash}m{0.13\textwidth}@{}}
\toprule
Benchmark & Prompt & {GPT-4o} & {Claude 3.5~S.} & {Gemini 1.5~P.} \\ 
\toprule
\multirow{2}{*}{\makecell[l]{MMLU (Undergraduate \\Level Knowledge)}} & Zero Shot & \textbf{88.70\%} & 88.30\% & - \\
& Five Shot & - & \textbf{88.70\%} & 85.90\% \\ 
\midrule
GPQA (Graduate Level Reasoning) & Chain of Thought & 53.60\% & \textbf{59.40\%} & 46.20\% \\ 
\midrule
DROP (Reasoning) & Three Shot & 83.40\% & \textbf{87.10\%} & 74.90\% \\
\bottomrule
\end{tabular}
\end{table*}

%-----------------------------------------------------
%-----------------------------------------------------
\section*{Results}
\label{Sec:results}
%Subheadings should be used

\subsection*{Accuracy}

\subsubsection*{Overall Performance}

The results show that all the approaches obtained a score of over 88\% in selecting the correct option for the 1050 RD exam questions, as indicated in Figure \ref{fig:accuracy_percentage} and Table \ref{fig:accuracy_percentage}. Overall, GPT-4o achieved the highest score (the blue markers in the figure) ranging between 91\% and 95\%, with the best score for CoT-SC. On the other hand, Gemini 1.5 Pro (the green markers) had the lowest scores.

In both GPT-4o and Claude 3.5 Sonnet, the CoT and CoT-SC prompting techniques resulted in similar percentage scores, which were approximately 2.5 percent higher than the ZS prompting's scores. However, the combination of Gemini with CoT or CoT-SC did not improve the accuracy but produced wider percentage scores across repeated measurements, with ranges of 1.9 and 1.2. Moreover, RAP obtained better scores, compared to ZS, in GPT-4o but slightly decreased the performance of Claude and Gemini models.

In addition to our findings, an overview of the models' performance on existing knowledge and reasoning benchmarks are indicated in Table \ref{table:accuracy_percentage_other_benchmarks}. The performance scores of these three benchmarks were collected from \cite{anthropic2024claude,openai2024gpt4,reid2024gemini}. The GPQA benchmark \cite{rein2023gpqa} includes 448 multiple-choice questions on biology, physics, and chemistry. The MMLU benchmark \cite{hendrycks2020measuring} contains multiple-choice questions from 57 topics, such as elementary mathematics, US history, computer science, and law; and the DROP benchmark \cite{dua2019drop} consists of 96,567 questions focusing on discrete reasoning over the content of paragraphs, including addition, counting, and sorting. Claude 3.5 Sonnet outperformed the other LLMs in all scenarios, except for MMLU using the ZS prompting. These findings do not fully align with our findings presented in Table \ref{table:accuracy_percentage}.

\begin{figure*}[!t]
    \centering
    \begin{subfigure}[t]{0.88\textwidth}
        \centering
        \includegraphics[width=\textwidth]{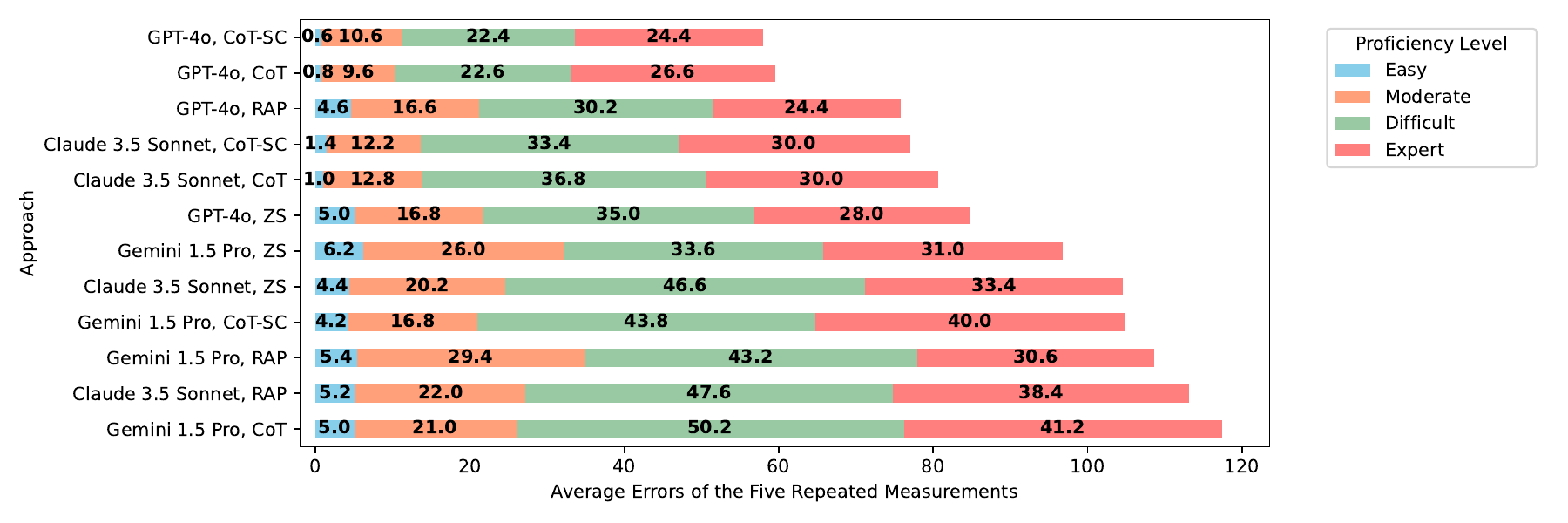}
        \caption{Average errors per approach by proficiency level. The exam includes 149 Easy, 352 Moderate, 392 Difficult, and 157 Expert levels questions.}
        \label{fig:error_subgroup_difficulty}
    \end{subfigure}
    \hfill
    \begin{subfigure}[t]{0.88\textwidth}
        \centering
        \includegraphics[width=\textwidth]{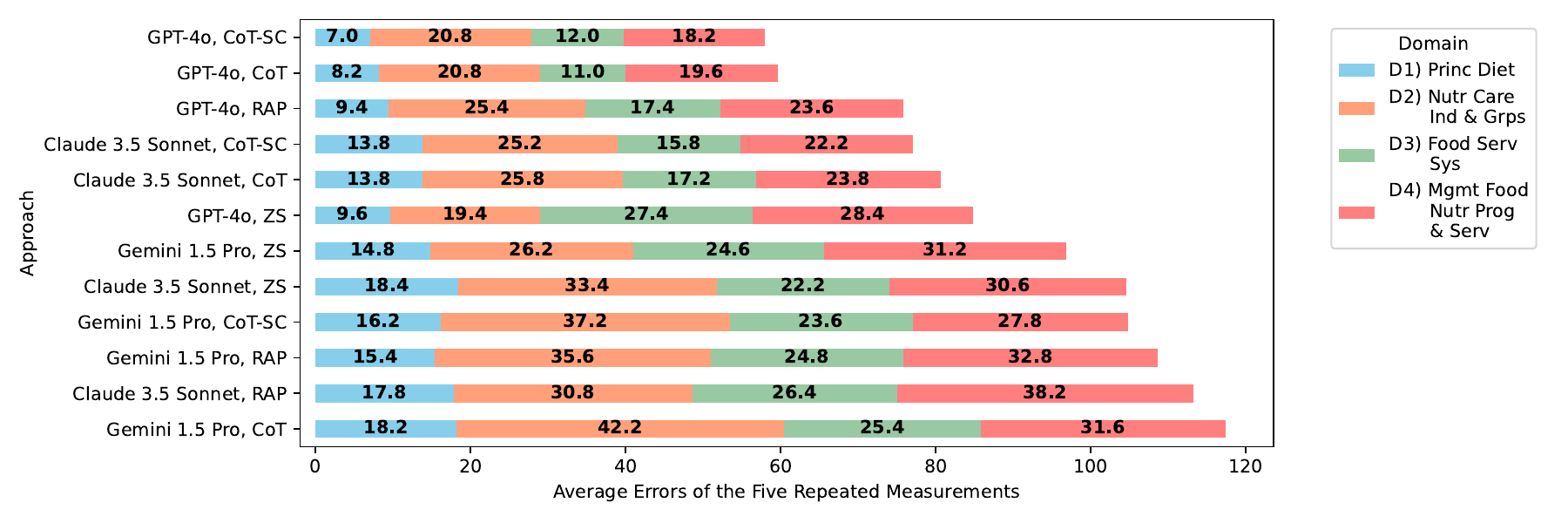}
        \caption{Average errors per approach by domain. The exam includes 237 principles of dietetics, 392 nutrition care for individuals and groups, 185 food service systems, and 236 management of food and nutrition programs and services questions.}
        \label{fig:error_subgroup_domain}
    \end{subfigure}
    \caption{The LLMs' inaccurate responses based on the RD exam questions' proficiency levels and domains.}
    \label{fig:error_subgroup}
\end{figure*}

\subsubsection*{Subgroup Error Analysis}

We categorize the RD exam questions into different subgroups, within which the LLMs' inaccurate responses are assessed. To achieve this, we analyze the errors obtained in terms of proficiency levels and four nutrition domains (i.e., topics).

\textbf{Proficiency Levels:}
The approaches are evaluated based on the questions' proficiency levels, provided by the Academy of Nutrition and Dietetics, eatrightPREP for the RDN Exam \cite{academy2024rdexam}. The exam consists of 149 Easy, 352 Moderate, 392 Difficult, and 149 Expert levels questions. Figure \ref{fig:error_subgroup_difficulty} shows the average errors for each approach. 

GPT-4o obtained the lowest overall average error counts. The model with CoT-SC resulted in the fewest errors across the proficiency levels, with the average errors of 0.6, 10.6, 22.4, and 24.4 for Easy, Moderate, Difficult, and Expert levels questions, respectively. Compared to ZS prompting, CoT and CoT-SC improved the model's performance at all levels, but RAP only enhanced the responses of the Difficult and Expert level questions.

Similar to the GPT-4o approaches, Claude~3.5 Sonnet performance was enhanced by CoT and CoT-SC. Claude 3.5 Sonnet with CoT and CoT-SC achieved similar average error rates. Conversely, using Claude 3.5 Sonnet, RAP recorded the highest error counts, particularly with 5 more errors (on average) for Expert questions, compared to the ZS prompting technique. 

Gemini 1.5 Pro had the highest number of errors overall. The ZS prompting recorded the lowest average errors with Gemini. Compared to ZS, CoT and CoT-SC improved the responses of the Moderate questions but obtained higher average errors for the Difficult and Expert level questions. RAP obtained higher error rates for Moderate and Difficult questions.

\begin{figure*}[t!]
    \centering
    \includegraphics[width=0.7\textwidth]{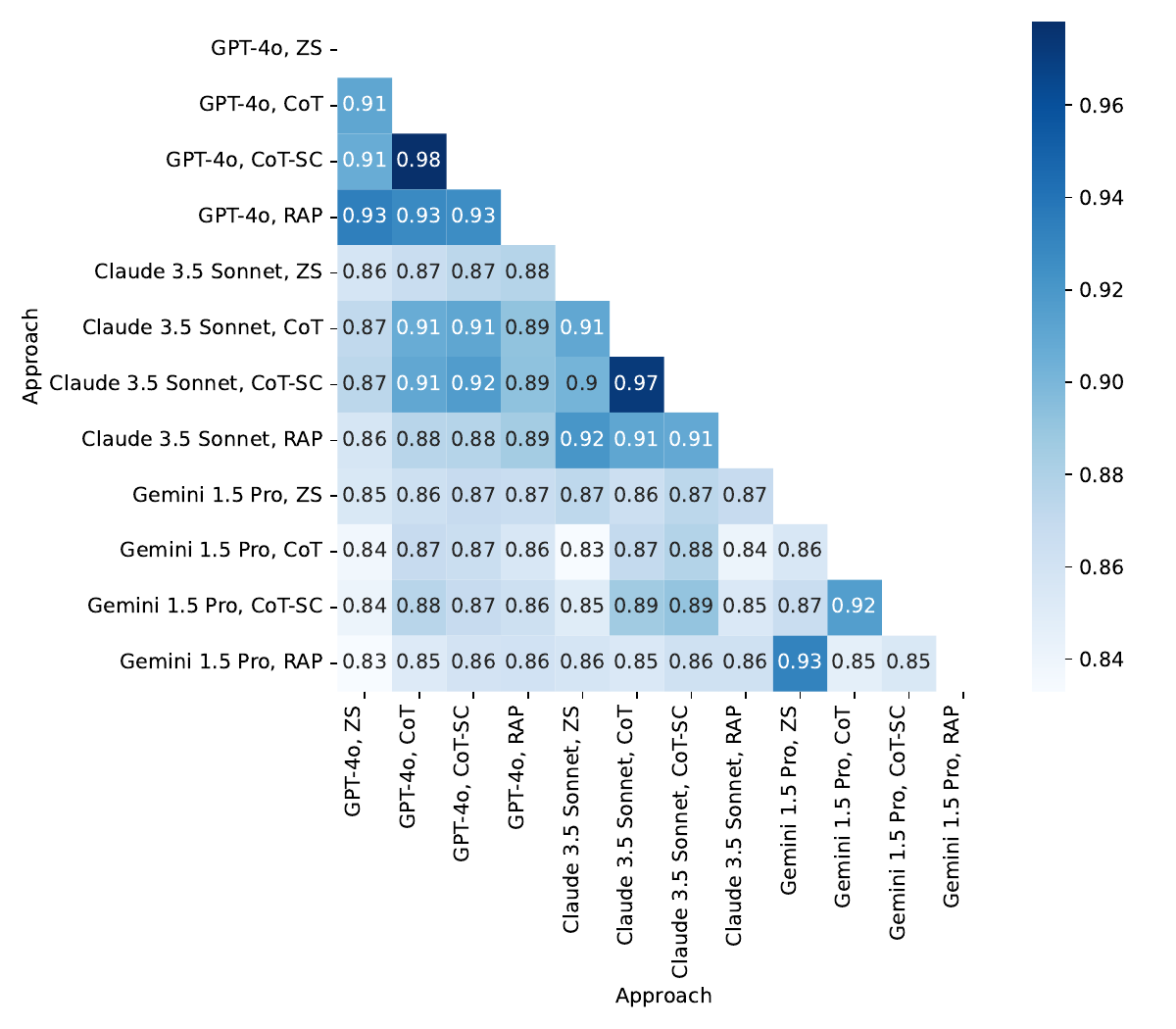}
    \caption{The Cohen's Kappa coefficients measured for each of the 12 pairwise comparisons using the RD exam. The dark blue indicates high levels of agreement, while the light blue represents lower agreement levels.}
    \label{fig:cohens_kappa}
\end{figure*}

\textbf{Domains:}
The inaccurate responses collected by each approach is evaluated based on four domains: \textit{D1) Principles of Dietetics}, \textit{D2) Nutrition Care for Individuals and Groups}, \textit{D3) Food Service System}, and \textit{D4) Management of
Food and Nutrition Programs and Service}. The exam consists of 237, 392, 185, and 236 questions for D1, D2, D3, and D4, respectively. As illustrated in Figure \ref{fig:error_subgroup_domain}, the impact of prompt engineering techniques varied across the domains for the three LLMs.

GPT-4o with CoT-SC reduced the average error counts in D3 from 27.4 to 12 and in D4 from 28.4 to 18.2, compared to GPT-4o with ZS. CoT and RAP also showed similar improvements in error rates although RAP recorded more errors for D2. Using GPT-4o, different prompting techniques resulted in small changes in the error rates observed in D1. 

Claude 3.5 Sonnet showed that transitioning from ZS prompting to CoT-SC or CoT reduced the average errors across the four domains. On the other hand, RAP slightly improved D1 and D2 but obtained more errors in D3 and D4. 

With Gemini 1.5 Pro, different prompts led to small variations in error counts, with changes of fewer than 4 errors on average in D1, D3, and D4. However, ZS prompting obtained the lowest error count in D2, with an average of 26.2 errors. Nevertheless, this outcome shows approximately 6 errors higher than the performance achieved by GPT-4o. In D2, Gemini and CoT obtained the highest error rates.

\subsection*{Consistency}

\subsubsection*{Inter-rater Analysis}
The inter-rater reliability of the responses from the approaches was analyzed to investigate their agreement. To achieve this, Cohen's Kappa coefficient was calculated for each pair of approaches to determine if they selected the same choices, whether accurate or inaccurate. Our study includes 12 distinct approaches (3 LLMs multiplied by 4 prompting techniques), so Cohen's Kappa was measured for each of the 12 pairwise comparisons. Since each approach is repeated five times, one set of measurements per approach is randomly selected to assess the inter-rater reliability. Figure~\ref{fig:cohens_kappa} presents the Cohen's Kappa coefficients, where dark blue indicates high levels of agreement, and light blue represents lower agreement levels. Additionally, the detailed statistical data are presented in Supplementary Table~S.1.

The approaches based on GPT-4o showed a high degree of agreement, indicated by a Cohen's Kappa coefficient of 0.98 between CoT and CoT-SC and a coefficient of 0.93 between RAP and the other three prompting techniques. This confirms that altering these prompting techniques did not result in a substantial change in the GPT-4o's behavior. Similarly, Claude 3.5-based approaches indicated comparable levels of agreement. In contrast, the Gemini 1.5 Pro's approaches recorded relatively lower Cohen's Kappa coefficients, despite maintaining high overall agreement. The Cohen's Kappa coefficients of the Gemini-based approaches were from 0.84 to 0.93. The agreement level between CoT and CoT-SC was 0.92, and the agreement between ZS and RAP was 0.93. Interestingly, among the prompting techniques, the approaches (even with different LLMs) using CoT and CoT-SC obtained higher levels of agreement.

\subsubsection*{Intra-rater Analysis}

In this study, each approach was repeated five times, resulting in five sets of responses. The intra-rater reliability of the responses was evaluated by measuring the repeatability of the approaches, determining how consistently they agreed with themselves when receiving the same questions. For this purpose, Fleiss Kappa was employed to assess the intra-rater agreements. Table~\ref{table:fleiss_kappa} indicates the Fleiss Kappa coefficients, and Supplementary Table~S.2 includes the detailed statistical data. 

Gemini 1.5 Pro combined with the ZS prompting achieved the highest agreement among all combinations, whereas the Gemini with CoT produced the lowest agreement. The approaches based on Claude 3.5 Sonnet demonstrated the highest overall agreement. For the three LLMs, the ZS prompting technique consistently resulted in the highest agreement, as indicated by Fleiss's Kappa coefficients of 0.996, 0.987, and 0.980 for Gemini 1.5 Pro, Claude 3.5 Sonnet, and GPT-4o, respectively. Similarly, the coefficients of the LLMs with RAP were high. The CoT-SC recorded the third highest agreement, while the CoT obtained the lowest.

\begin{table}[!t]
\small
\caption{The Fleiss Kappa coefficients of the 12 approaches. Each approach was repeated 5 times.}
\label{table:fleiss_kappa}
\centering
\begin{tabular}{@{}>{\arraybackslash}m{0.20\columnwidth}>{\arraybackslash}m{0.15\columnwidth}>{\centering\arraybackslash}m{0.18\columnwidth}>{\centering\arraybackslash}m{0.25\columnwidth}@{}}

\toprule
                    &    & \multicolumn{1}{c}{Fleiss' Kappa} & \multicolumn{1}{c}{95\% CI} \\ \midrule
\multirow{4}{*}{GPT-4o} & ZS & 0.980 & 0.973 -- 0.987 \\
                    & CoT & 0.969 & 0.960 -- 0.977 \\
                    & CoT-SC & 0.977 & 0.970 -- 0.985   \\
                    & RAP & 0.985 & 0.978 -- 0.991     \\
\midrule
\multirow{4}{*}{Claude 3.5 S.} &  ZS & 0.987 & 0.981 -- 0.992 \\
                    & CoT & 0.975 & 0.967 -- 0.983 \\
                    & CoT-SC & 0.982 & 0.975 -- 0.988 \\
                    & RAP & 0.977 & 0.970 -- 0.985 \\
\midrule
\multirow{4}{*}{Gemini 1.5 P.} & ZS & 0.996 & 0.993 -- 0.999 \\
                    & CoT & 0.902 & 0.887 -- 0.917 \\
                    & CoT-SC & 0.938 & 0.926 -- 0.951 \\
                    & RAP & 0.991 & 0.987 -- 0.996 \\ \midrule
\end{tabular}
\end{table}

% -----------------------------------------------------
% -----------------------------------------------------
\section*{Discussion}
\label{Sec:discussion}
% No subheadings, limitations, or conclusions sections are permitted

Our findings indicated that all the approaches, combining three LLMs with four prompt engineering techniques, successfully passed the RD exam. However, the three leading LLMs had different performance levels in terms of the number of inaccurate responses and consistency. In addition, the prompting techniques had considerable impacts on the results. Such prompting impacts were also explored in other evaluation studies, for example, in clinical medicine \cite{wang2024prompt}, mental health \cite{grabb2023impact} and radiology \cite{russe2024improving}. 

The combination of GPT-4o with CoT-SC prompting outperformed the other approaches in terms of accuracy, while Gemini 1.5 Pro with ZS prompting showed the highest consistency. On the other hand, the lowest average percentage score was 89.22\% for Gemini 1.5 Pro with CoT, which also showed the lowest agreement in repeated measurements, with a coefficient of 0.902. GPT-4o recorded the highest accuracy overall (see~Table \ref{table:accuracy_percentage}). 

This outcome contrasts with previous non-nutrition research, except in MMLU \cite{hendrycks2020measuring} with ZS prompting (see Table~\ref{table:accuracy_percentage_other_benchmarks}). Claude 3.5 with CoT obtained a 59.4\% score on GPQA \cite{rein2023gpqa}. However, the three LLMs using CoT on the RD exam achieved scores above 90\%. This difference might be due to the different difficulty levels of the exams. Particularly, 14.9\% of the questions in the RD Exam are at the Expert level. However, as reported by Rein \textit{et al.} \cite{rein2023gpqa}, the GPQA questions are \textit{``extremely difficult,''} from which PhD students achieved a 65\% score while non-expert individuals achieved a 34\% score. Moreover, DROP \cite{dua2019drop} demonstrated that Claude 3.5 with Three Shot prompting outperformed in reasoning over text. Conversely, our results indicated that GPT-4o performed better using the reasoning process of CoT prompting.

Prior nutrition-focused research indicated that ChatGPT was accurate in most nutrition instances, but the chatbot also recorded errors that could potentially harm and negatively impact the end-users. Therefore, achieving general accuracy is insufficient for practical real-world applications. For example, Sun \textit{et al.}~\cite{sun2023ai} indicated that ChatGPT-3.5 and ChatGPT-4 passed the Chinese RD exam (included 200 questions) and the food recommendations were acceptable despite the presence of mistakes for specific foods, such as root vegetables and dry beans. Mishra \textit{et al.}~\cite{mishra2024evaluation} tested ChatGPT in eight medical nutritional therapy scenarios and discussed that ChatGPT should be avoided for complex scenarios. Similarly, other studies~\cite{ponzo2024chatgpt,pugliese2024accuracy} discussed that ChatGPT has great potentials for nutritional management focusing on non-communicable diseases, but the model might be potentially harmful by providing inaccurate responses, particularly in complex situations. Another study~\cite{hoang2023consistency} leveraged ChatGPT-3.5 and ChatGPT-4 to provide nutritional information for eight menus. Their results indicated that responses had no significant differences compared to nutritionists' recommendations in terms of energy, carbohydrate, and fat contents, but the difference was statistically significant for protein. The potential of ChatGPT to generate dietary advice for patients with allergic to food allergens were also investigated \cite{niszczota2023credibility}. It was shown that although the model was generally accurate, it produced harmful diets. These studies highlight the need for further investigation into LLM responses within the context of food and nutrition.

Our results confirmed previous findings about the overall accuracy of ChatGPT and the instances of inaccurate responses. However, unlike the existing work, our study is not merely restricted to ChatGPT or the ZS prompting technique. We focused on examining errors across various subcategories and mitigate them by employing prompting techniques (reasoning and ensemble) and external knowledge retrieval.

\textbf{CoT} guided LLMs to perform a reasoning process when answering a question. Our findings showed that CoT, compared to ZS prompting, enhanced the accuracy of GPT-4o and Claude 3.5 Sonnet but led to diminished consistency. The LLMs with CoT do not consistently generate the same reasoning paths, even with identical prompts (see Table~\ref{table:fleiss_kappa}). This variability indicates randomness in the selection of reasoning paths.

We observed that the reasoning steps of CoT considerably reduced the LLMs' mistakes for the questions with Easy, Moderate, and Difficult proficiency levels, but this improvement was less for Expert-level questions, where only a few errors were corrected. Additionally, CoT notably improved the questions about \textit{D3) food service systems}, which involved calculations for food cost and portion estimation/forecasting. CoT also enhanced the accuracy of \textit{D4) food and nutrition management}, which included theoretical and conceptual questions requiring an understanding of implicitly stated relationships. These improvements by CoT are consistent with existing literature, indicating CoT enhances LLMs' performance in arithmetic and commonsense tasks by establishing logical connections \cite{wei2022chain}. Conversely, the combination of Gemini 1.5 Pro with CoT showed different patterns, where both accuracy and consistency decreased. Gemini with CoT was unable to select a choice from the given multiple-choice options for 20 out of 1050 questions (on average). Although the errors on Easy and Moderate levels questions slightly decreased, the errors on Difficult and Expert levels questions notably increased.

It should be noted that while CoT reduced errors in questions requiring calculations, our observations indicate that CoT responses still include miscalculations and rounding errors. This issue may arise due to the inherent characteristics of Transformer models, designed to generate text based on tokens rather than numerical values. Potential solutions to address these issues include agentic approaches \cite{gou2023tora,abbasian2023conversational}, which integrate LLMs with calculator tools or symbolic computing systems.

\textbf{CoT-SC} guided LLMs to perform multiple independent reasoning processes, then the responses were merged using a majority voting method. Our findings revealed that CoT-SC (compared to CoT) improved accuracy, particularly in Gemini 1.5 Pro. However, in GPT-4o and Claude 3.5, this improvement was small, as it only led to the correction of a few errors. This small difference can also be observed in their high inter-rater coefficient agreement, as illustrated in Figure \ref{fig:cohens_kappa}. This finding does not support the literature suggesting that CoT-SC considerably enhances the accuracy of CoT \cite{wang2022self}. 

On the other hand, CoT-SC achieved notably higher consistency (intra-rater agreement) compared to CoT. The ensemble process enabled by CoT-SC mitigates the randomness in the selection of reasoning paths. For GPT-4o and Claude 3.5 Sonnet, the Fleiss' Kappa agreements of CoT-SC were as robust as the agreements of ZS prompting. The Gemini's inability to select a choice from the given multiple-choice options also improved, reducing them from 20 in CoT to 6 in CoT-SC. This highlights the importance of employing such ensemble techniques to enhance the consistency of LLM's reasoning process by combining multiple reasoning paths rather than relying on a single path.

\textbf{RAP} integrated external relevant information from multiple references into the input prompts. However, our findings showed that RAP did not consistently improve accuracy across the three models. GPT-4o effectively leveraged the retrieved information to reduce error rates, particularly for Difficult and Expert questions that required more comprehensive understanding. Similar to CoT and CoT-SC, RAP improved \textit{D3) food service systems} and \textit{D4) food and nutrition management} questions. Although relevant information was provided in our knowledge base, RAP (compared to ZS) has recorded higher error rates for \textit{D2) nutrition care}. D2 questions are mostly related to medical nutrition therapy, dietary guidelines, counseling skills, and nutrition care process. This higher error rates might arise from irrelevant retrieval, where the retrieval model fetches extraneous information \cite{gao2023retrieval}. Additionally, the complexity or ambiguity of the queries might contribute to this problem making it challenging for the retrieval model to find the most relevant chunks.

In contrast to GPT-4o, Gemini 1.5 Pro with RAP showed opposite behavior, as the accuracy for the Difficult and Expert questions reduced. We noticed that, in some cases, Gemini was prioritizing external information over its own internal knowledge, even when that external information was irrelevant to the question. This resulted in incorrect interpretations and answers. For example, for two questions, the model generated \textit{``The provided text does not contain the answer to the question as it pertains to dietary restrictions for patients on Linezolid.''} and \textit{``The provided text focuses on Body Mass Index (BMI) but does not contain information about when weight and BMI peak.''} This issue was particularly observed in D2, where error rates increased from 26.2 (ZS) to 35.6 (RAP).

It is worth noting that the prompting techniques had less impact, whether positive or negative, on \textit{D1) Principles of Dietetics} questions compared to the other domains. D1 questions primarily focus on general food science, nutrients, biochemistry, and related research (e.g. \textit{which fruit has the highest fructose?}), compared to the other domains that are more specialized in dietetics or involve more domain knowledge. For D1, GPT-4o achieved the best accuracy.  

This study is limited to the leading proprietary LLM models. These models are user-friendly and highly powerful. Our results also confirm their significant potential in food and nutrition applications. Yet, growing concerns are being raised about their lack of openness and limited access. In contrast, open-source LLMs are emerging rapidly, offering benefits, such as improved data security and privacy, decreased reliance on vendors, and the ability to customize models. Examples of the state-of-the-art open-source LLMs are Llama 3 \cite{metaai2024llama3}, Falcon 2 \cite{tii2024falcon2}, and Yi-34B \cite{01ai2024yi34b}. Given their advantages, future research should evaluate the performance of open-source LLMs in the diet and nutrition field.

Our evaluation has primarily concentrated on the accuracy and consistency of the models. Given the sensitivity of health and nutrition applications, ensuring high accuracy and consistency is essential. However, it is important to assess LLMs from other perspectives, such as safety, bias, privacy, and emotional support, to mention a few \cite{abbasian2024foundation,liang2022holistic,sun2024trustllm}. Future work in this direction will involve evaluating LLMs according to these trustworthiness metrics by leveraging patient-centric questions, answers, and conversations. 

Additionally, we examined the impacts of prompt engineering methods on LLM answers to diet and nutrition questions. Various studies have explored the role of fine-tuning \cite{xu2023parameter,singhal2023large,zhang2024comparison} and agentic methods \cite{yang2024chatdiet,abbasian2024knowledge,li2024personal}. Future research should evaluate their impact on nutrition management applications.

In conclusion, this study assessed the accuracy and consistency of the GPT-4o, Claude 3.5 Sonnet, and Gemini 1.5 Pro in responding to diet and nutrition questions of the RD exam. In contrast to the previous LLM evaluation studies focusing on nutritional management, our experiments were not restricted to ChatGPT or ZS prompting. We evaluated the models using the RD exam and analyzed their errors across various questions complexities and nutrition domains. Our findings highlighted the strengths and weaknesses of the three LLMs, showing the influence of different prompting techniques on their responses to the RD exam questions. GPT-4o with CoT-SC prompting outperformed other approaches, while Gemini 1.5 Pro with ZS indicated the highest consistency. For GPT-4o and Claude 3.5, the application of CoT improved accuracy, while CoT-SC enhanced both accuracy and consistency. RAP particularly improved GPT-4o performance in addressing difficult- expert-level questions. Consequently, selecting the appropriate LLM and prompt engineering, tailored to the proficiency level and specific domain, can considerably reduce errors and mitigate potential risks in diet and nutrition chatbot applications.

% -----------------------------------------------------
% -----------------------------------------------------
\section*{Methods}
\label{Sec:methods}
% Subheadings should be used. For clinical trials, it is mandatory to provide the trial registration number and date of registration. If the Article reports results from a randomized clinical trial, it is mandatory to include a completed CONSORT checklist or the appropriate extension checklist in the Supplementary Information. If your study includes human or animal participants, human or animal material, or human data, it is mandatory to provide details of the ethical approval(s) obtained and how human participants consented to take part in the study. If the Article reports a pre-clinical animal study, authors are strongly encouraged to include a completed ARRIVE Essential 10 checklist in the Supplementary Information. All descriptions of Methods should be provided in the main manuscript file and not in the Supplementary Information

\begin{figure*}[bp]
    \centering
    \begin{subfigure}[t]{0.24\textwidth}
        \centering
        \includegraphics[width=\textwidth]{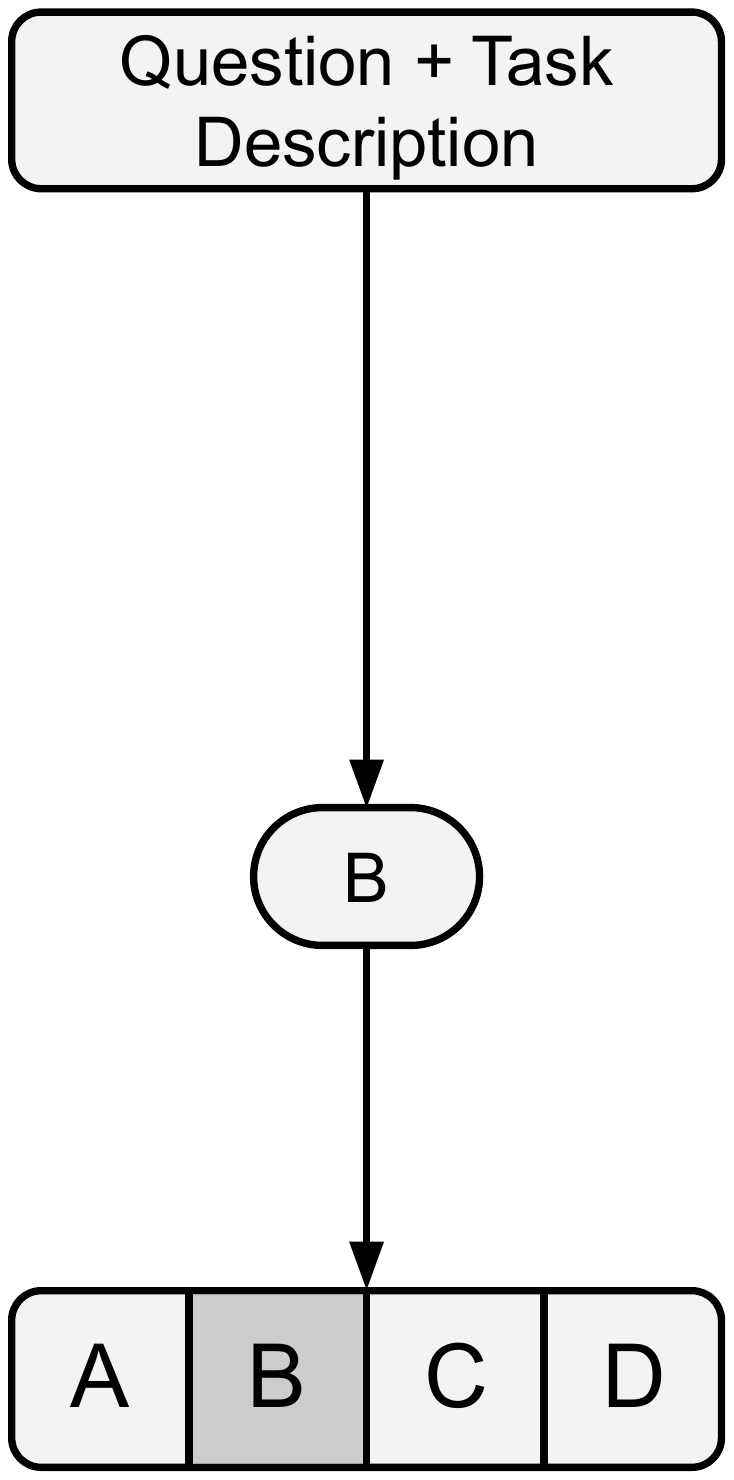}
        \caption{Zero Shot (ZS) prompting}
        \label{fig:ZS}
    \end{subfigure}
    \hfill
    \begin{subfigure}[t]{0.24\textwidth}
        \centering
        \includegraphics[width=\textwidth]{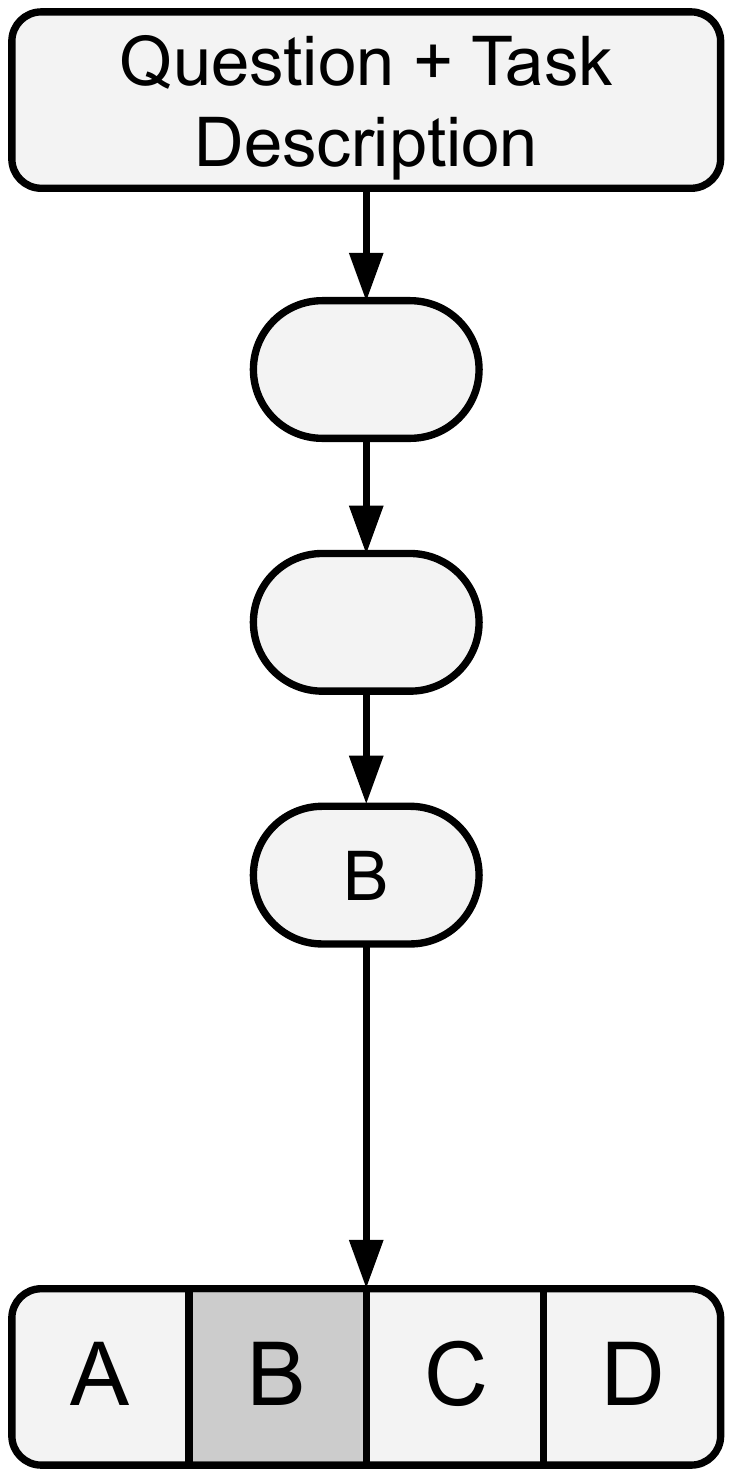}
        \caption{Chain of Thought (CoT) prompting}
        \label{fig:CoT}
    \end{subfigure}
    \hfill
    \begin{subfigure}[t]{0.24\textwidth}
        \centering
        \includegraphics[width=\textwidth]{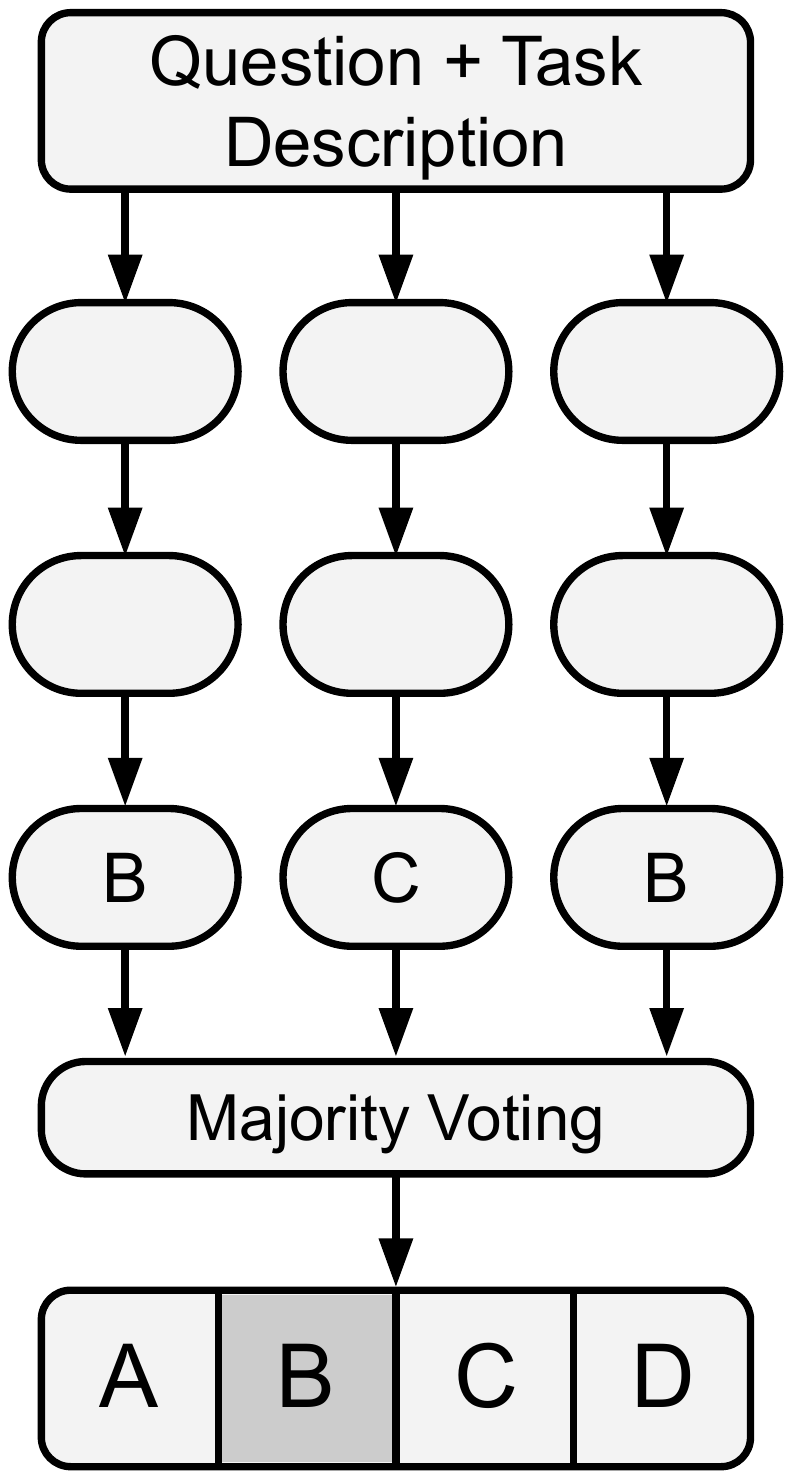}
        \caption{Chain of Thought with Self Consistency (CoT-SC) prompting}
        \label{fig:CoT-SC}
    \end{subfigure}
    \hfill
    \begin{subfigure}[t]{0.24\textwidth}
        \centering
        \includegraphics[width=\textwidth]{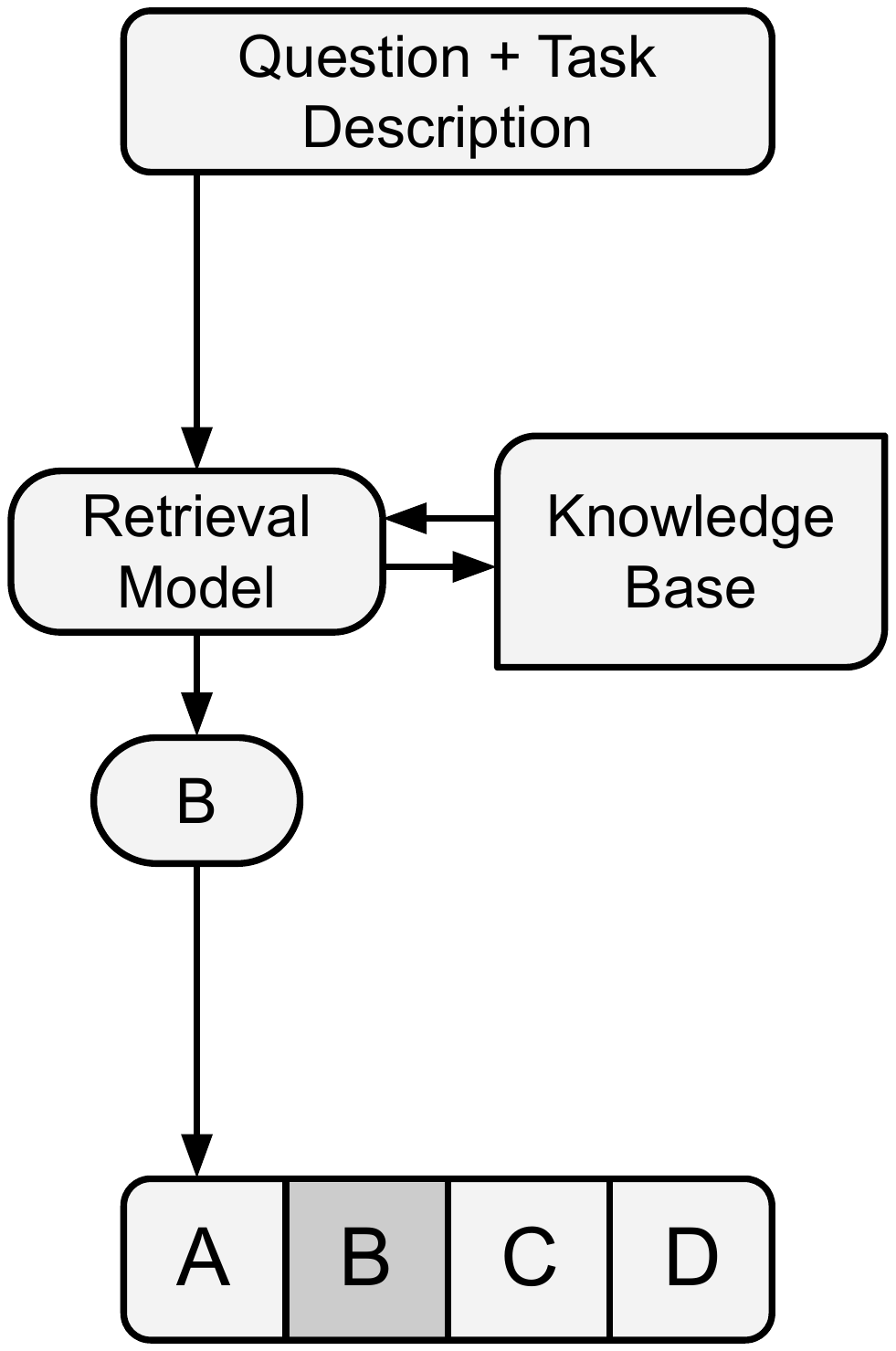}
        \caption{Retrieval Augmented Prompting (RAP)}
        \label{fig:RAP}
    \end{subfigure}
    
    \caption{Schematic illustrations of the four prompting techniques used in the evaluation. The inputs include multiple-choice questions along with task description, and the generated output includes the selected choice.}
    \label{fig:prompt_engineering_techniques}
\end{figure*}

\subsection*{Registered Dietitian Exam}

The Registration Examination for Dietitians is a required exam for individuals seeking to obtain the registered dietitian credential. To take the exam, candidates must successfully complete the eligibility requirements provided by the Commission on Dietetic Registration (CDR) \cite{cdr}. The examination is computer-based and consists of 125 to 145 four-choice questions \cite{cdr2024rdexamhandbook}. The exam includes multiple-choice questions from four major domains: D1) Principles of Dietetics (21\%), D2) Nutrition Care for Individuals and Groups (45\%), D3) Food Service Systems (13\%), and D4) Management of Food and Nutrition Programs and Services (21\%) \cite{cdr2024rdexamhandbook}. The exam is scored from 1 to 50, and the minimum score to pass is 25. The score is calculated based on the candidate’s performance as well as the difficulty levels of the questions \cite{cdr2024rdexamhandbook}. 

Within the four domains, D1 covers topics related to i) food, nutrition, and supporting sciences, ii) education, communication and technology, and iii) research applications. D2 consists of the topics related to i) screening and assessment, ii) diagnosis, iii) planning and intervention, and iv) monitoring and evaluation. D3 includes topics related to i) menu development, ii) procurement, production, distribution, and service, iii) sanitation and safety, and iv) equipment and facility planning. D4 includes topics related to i) functions of management, ii) human resource management, iii) financial management, iv) marketing and public relations; and v) quality management and regulatory compliance \cite{cdr2024rdexamhandbook}.

\subsection*{Large Language Models}
GPT-4o, Claude 3.5 Sonnet, and Gemini 1.5 Pro, as the leading LLMs chatbots \cite{leaderboardchatbotarena,leaderboardartificialanalysis}, are employed in this study for evaluation. OpenAI released GPT-4o, their new flagship model, on May 13, 2024 \cite{openai2024gpt4}, Claude 3.5 Sonnet was launched, by Anthropic, as their strongest vision model yet, on Jun 20, 2024 \cite{anthropic2024claude}, and Google announced Gemini 1.5 Pro as their next-generation model on February 15, 2024 \cite{google2024gemini}. An overview of the models' performance on other benchmarks are indicated in Table \ref{table:accuracy_percentage_other_benchmarks}. Find more details in \cite{anthropic2024claude,openai2024gpt4,reid2024gemini}.

In this study, we set the temperature setting to 0 for all the models to better evaluate the LLMs' knowledge and decision-making in nutrition and diet applications, minimizing the effect of external variables on consistency. The temperature parameter, ranging from 0 to 2, regulates the uncertainty or randomness in the output \cite{peeperkorn2024temperature}. With a temperature setting of 0, the model generates responses by selecting the next words with the highest probability, making the model ``more deterministic.'' 

\subsection*{Prompt Engineering}
Four prompting techniques are utilized in this study for the models evaluation. Schematic illustrations of the four techniques are shown in Figure~\ref{fig:prompt_engineering_techniques}. Additionally, the instructions used for the prompting techniques are presented in Supplementary Table~S.3.

\textit{1) Zero Shot (ZS) prompting} generates the simplest type of prompt, including a question and a fixed task description. The model leverages its internal knowledge to generate responses \cite{sahoo2024systematic}. To the best of our knowledge, existing evaluations of LLM chatbots focusing on nutrition and diet have utilized ZS prompting for their assessments. \textit{2) Chain of Thought (CoT) prompting} consists of a question and a description to the model to answer the question through intermediate reasoning steps \cite{wei2022chain}. CoT has been widely used in medical studies \cite{holmes2023evaluating,wang2024prompt}. \textit{3) Chain of Thought with Self Consistency (CoT-SC)} prompting creates several independent reasoning paths using CoT. Subsequently, the outcomes are aggregated \cite{wang2022self}. In our experiments, we selected three independent reasoning paths and used a majority voting method for the aggregation. \textit{4) Retrieval Augmented Prompting (RAP)} fetches relevant information from a knowledge base in real-time and integrates it into the input prompt \cite{gao2023retrieval,li2023chatdoctor}. In contrast to the other prompting techniques, using RAP, the model generates responses by relying not only on its internal knowledge but also on external information. In our study, the knowledge base includes 125 documents (such as articles, books, and guidelines)  recommended by the Academy of Nutrition and Dietetics \cite{academy2024rdexam}, as references for the RD exam. The full list of the references used for RAP is provided in Supplementary Table~S.4. For the implementation, we leveraged a conventional Retrieval Augmented Generation (RAG) framework \cite{gao2023retrieval}. To achieve this, the references were divided into 512-token chunks, using the Amazon Titan Text Embeddings v2 model \cite{amazon2024titan} for text embeddings. Then, the Cosine Similarity method \cite{schutze2008introduction} was utilized to identify the most similar chunks.

\subsection*{Data Collection}

The 1050 RD exam questions were delivered to the three models using the four prompting techniques. Each question was asked five times. Consequently, we collected 60 (i.e., $3 \times 4 \times 5$) sets of 1050 responses. As previously mentioned, the questions include four choices. We observed that sometimes the LLMs were unable to select an option from the multiple choices and provided responses such as, \textit{``None of the above,"} \textit{``Since no option is correct, we cannot provide a final answer within the requested tags,''} or \textit{``Cannot be determined with the given information.''} In summary, this issue occurred once for GPT-4o with CoT, once for GPT-4o with CoT-SC, 15 times for Claude 3.5 with RAP, 100 times for Gemini 1.5 with CoT, 30 times for Gemini 1.5 with CoT-SC, and 63 times for Gemini 1.5 with RAP. For these responses, we added another option, labeled ``Others.'' 

The collected responses were compared with the ground truth answers provided by the Academy of Nutrition and Dietetics, eatrightPREP \cite{academy2024rdexam}. It should be noted that we used a new chat session for each query to minimize bias in the evaluation caused by information leakage from other questions. The data collection was performed in Python using OpenAI \cite{pythonopenai}, google-generativeai \cite{pythongemini}, Boto3 \cite{pythonboto3}, and lxml.etree \cite{pythonlxml} libraries.

\subsection*{Statistical Analysis}

The responses were evaluated in terms of accuracy and consistency. Accuracy measures how close a set of responses are to the ground truth answers. To this end, we calculate the percentage score, which is the ratio of correct responses to all responses multiplied by 100. The percentage score indicates how well the LLMs can detect the correct option. As previously mentioned, each measurement is repeated five times. The five repeated measurements in each test are grouped, and the mean and standard deviation of the scores are calculated.

Consistency refers to the degree to which responses produce the same results. To assess consistency, we perform inter-rater and intra-rater analysis approaches. \cite{hallgren2012computing}. For the former, the agreement between the responses obtained from different models / prompting techniques are evaluated. To this end, Cohen's Kappa \cite{cohen1960coefficient} was utilized to measure the degree of agreement between two sets of responses. For example, the agreement between responses obtained from GPT-4o with ZS prompting and GPT-4o with CoT prompting are calculated. Furthermore, for the intra-rater analysis, Fleiss Kappa test \cite{fleiss1971measuring} was used to indicate the degree of overall agreement between the repeated measurements under fixed conditions. For instance, we assess whether GPT-4o with ZS prompting provides the same choices in repeated measurements. It should be noted that the statistical analysis was conducted in R Programming using irr \cite{irrrlib} and boot \cite{bootrlib} libraries.

% -----------------------------------------------------
% -----------------------------------------------------
\section*{Data Availability}
% Mandatory
The RD exam questions used in this study are not publicly available and can be accessed via \url{https://www.eatrightprep.org}.

% -----------------------------------------------------
% -----------------------------------------------------
\section*{Code Availability}
% Where applicable
The codes for data collection, API calls, and statistical analysis are available at
\url{https://github.com/iHealthLab/DietitianExamEval}.

% -----------------------------------------------------
% -----------------------------------------------------
%\section*{Acknowledgements}
% Where Funding for the study should be declared, a separate Funding statement is not permitted

% -----------------------------------------------------
% -----------------------------------------------------
%\section*{Author Contributions}
% Mandatory, the statement should refer to all authors individually, denoted by their initials

% -----------------------------------------------------
% -----------------------------------------------------
\section*{Competing Interests}
The authors declare no competing interests. Moreover, the funders of the study had no role in study design, data collection and analysis, or interpretation of results and preparation of the manuscript.
% Mandatory, you should declare all financial and non-financial competing interests for each author. If there are no competing interests to declare, this should be stated

%The authors declare no competing interests. Also, the funders of the study had no role in study design, data collection and analysis, or interpretation of results and preparation of the manuscript. \\

\bibliographystyle{naturemag}
{\footnotesize
\bibliography{main.bib}}
% Limited to 60 references, though not strictly enforced

\clearpage
\appendix
%\section*{Supplementary Materials}


\section*{Supplementary material}

Table S.1: A detailed report of Cohen's Kappa [1], showing the degree of agreement between each pair of approaches. It includes 12 approaches, consisting of 3 LLMs (i.e., GPT-4o [2], Claude 3.5 Sonnet [3], and Gemini 1.5 Pro [4]), and 4 prompting techniques (i.e., zero-shot (ZS), chain of thought (CoT), chain of thought with self consistency (CoT-SC), and retrieval augmented prompting (RAP)). All the P-values were approximately zero.

| Approach 1 | Approach 2 | Cohen's Kappa | 95% CI | Z value |
|---|---|---|---|---|
| GPT-4o, ZS | GPT-4o, CoT | 0.911 | $0.891 - 0.931$ | 51.1 |
| ” | GPT-4o, CoT-SC | 0.907 | $0.887 - 0.928$ | 50.9 |
| ” | GPT-4o, RAP | 0.934 | $0.916 - 0.952$ | 52.4 |
| ” | Claude 3.5 S., ZS | 0.858 | $0.833 - 0.883$ | 48.2 |
| ” | Claude 3.5 S. CoT | 0.872 | $0.848 - 0.895$ | 48.9 |
| ” | Claude 3.5 S., CoT-SC | 0.874 | $0.850 - 0.898$ | 49.0 |
| ” | Claude 3.5 S., RAP | 0.857 | $0.832 - 0.882$ | 48.2 |
| ” | Gemini 1.5 P., ZS | 0.855 | $0.831 - 0.881$ | 48.0 |
| ” | Gemini 1.5 P., CoT | 0.838 | $0.813 - 0.865$ | 47.8 |
| ” | Gemini 1.5 P., CoT-SC | 0.841 | $0.817 - 0.867$ | 47.4 |
| ” | Gemini 1.5 P., RAP | 0.833 | $0.806 - 0.860$ | 47.2 |
| GPT-4o, CoT | GPT-4o, CoT-SC | 0.978 | $0.968 - 0.989$ | 54.9 |
| ” | GPT-4o, RAP | 0.928 | $0.909 - 0.946$ | 52.0 |
| ” | Claude 3.5 S., ZS | 0.867 | $0.843 - 0.891$ | 48.6 |
| ” | Claude 3.5 S. CoT | 0.907 | $0.886 - 0.928$ | 50.9 |
| ” | Claude 3.5 S., CoT-SC | 0.910 | $0.889 - 0.930$ | 51.0 |
| ” | Claude 3.5 S., RAP | 0.876 | $0.852 - 0.898$ | 49.3 |
| ” | Gemini 1.5 P., ZS | 0.864 | $0.840 - 0.889$ | 48.5 |
| ” | Gemini 1.5 P., CoT | 0.869 | $0.845 - 0.893$ | 49.5 |
| ” | Gemini 1.5 P., CoT-SC | 0.876 | $0.853 - 0.899$ | 49.3 |
| ” | Gemini 1.5 P., RAP | 0.852 | $0.826 - 0.878$ | 48.2 |
| GPT-4o, CoT-SC | GPT-4o, RAP | 0.926 | $0.908 - 0.945$ | 52.0 |
| ” | Claude 3.5 S., ZS | 0.874 | $0.851 - 0.898$ | 49.1 |
| ” | Claude 3.5 S. CoT | 0.911 | $0.891 - 0.931$ | 51.1 |
| ” | Claude 3.5 S., CoT-SC | 0.917 | $0.897 - 0.937$ | 51.5 |
| ” | Claude 3.5 S., RAP | 0.877 | $0.854 - 0.900$ | 49.3 |
| ” | Gemini 1.5 P., ZS | 0.869 | $0.845 - 0.893$ | 48.8 |
| ” | Gemini 1.5 P., CoT | 0.866 | $0.843 - 0.890$ | 49.4 |
| ” | Gemini 1.5 P., CoT-SC | 0.869 | $0.845 - 0.894$ | 49.0 |
| ” | Gemini 1.5 P., RAP | 0.860 | $0.834 - 0.884$ | 48.7 |
| GPT-4o, RAP | Claude 3.5 S., ZS | 0.877 | $0.854 - 0.900$ | 49.2 |
| ” | Claude 3.5 S. CoT | 0.892 | $0.870 - 0.914$ | 50.0 |
| ” | Claude 3.5 S., CoT-SC | 0.893 | $0.872 - 0.916$ | 50.1 |
| ” | Claude 3.5 S., RAP | 0.885 | $0.861 - 0.908$ | 49.8 |
| ” | Gemini 1.5 P., ZS | 0.867 | $0.843 - 0.891$ | 48.6 |
| ” | Gemini 1.5 P., CoT | 0.856 | $0.832 - 0.881$ | 48.8 |
| ” | Gemini 1.5 P., CoT-SC | 0.864 | $0.841 - 0.888$ | 48.7 |
| ” | Gemini 1.5 P., RAP | 0.861 | $0.836 - 0.885$ | 48.7 |
| Claude 3.5 S., ZS | Claude 3.5 S. CoT | 0.910 | $0.889 - 0.930$ | 51.1 |
| ” | Claude 3.5 S., CoT-SC | 0.902 | $0.880 - 0.923$ | 51.7 |
| ” | Claude 3.5 S., RAP | 0.921 | $0.902 - 0.940$ | 51.8 |
| ” | Gemini 1.5 P., ZS | 0.873 | $0.850 - 0.895$ | 49.0 |
| ” | Gemini 1.5 P., CoT | 0.833 | $0.808 - 0.860$ | 47.6 |
| ” | Gemini 1.5 P., CoT-SC | 0.850 | $0.826 - 0.875$ | 48.0 |

Table S.1: A detailed report of Cohen's Kappa [1], showing the degree of agreement between each pair of approaches (continued).

| Approach 1 | Approach 2 | Cohen's Kappa | 95% CI | Z value |
|---|---|---|---|---|
| " | Gemini 1.5 P., RAP | 0.858 | 0.833 − 0.882 | 48.7 |
| Claude 3.5 S., CoT | Claude 3.5 S. CoT-SC | 0.972 | 0.960 − 0.984 | 54.5 |
| " | Claude 3.5 S., RAP | 0.911 | 0.891 − 0.931 | 51.3 |
| " | Gemini 1.5 P., ZS | 0.865 | 0.841 − 0.890 | 48.5 |
| " | Gemini 1.5 P., CoT | 0.870 | 0.845 − 0.893 | 49.6 |
| " | Gemini 1.5 P., CoT-SC | 0.886 | 0.864 − 0.909 | 49.9 |
| " | Gemini 1.5 P., RAP | 0.854 | 0.830 − 0.880 | 48.4 |
| Claude 3.5 S., CoT-SC | Claude 3.5 S. RAP | 0.909 | 0.888 − 0.930 | 51.1 |
| " | Gemini 1.5 P., ZS | 0.874 | 0.851 − 0.898 | 49.0 |
| " | Gemini 1.5 P., CoT | 0.878 | 0.854 − 0.901 | 50.0 |
| " | Gemini 1.5 P., CoT-SC | 0.891 | 0.868 − 0.913 | 50.2 |
| " | Gemini 1.5 P., RAP | 0.863 | 0.839 − 0.888 | 48.9 |
| Claude 3.5 S., RAP | Gemini 1.5 P., ZS | 0.869 | 0.846 − 0.894 | 48.9 |
| " | Gemini 1.5 P., CoT | 0.841 | 0.815 − 0.867 | 48.1 |
| " | Gemini 1.5 P., CoT-SC | 0.854 | 0.829 − 0.879 | 48.3 |
| " | Gemini 1.5 P., RAP | 0.863 | 0.839 − 0.888 | 49.1 |
| Gemini 1.5 P., ZS | Gemini 1.5 P., CoT | 0.856 | 0.831 − 0.881 | 48.8 |
| " | Gemini 1.5 P., CoT-SC | 0.867 | 0.843 − 0.891 | 48.8 |
| " | Gemini 1.5 P., RAP | 0.932 | 0.914 − 0.950 | 52.8 |
| Gemini 1.5 P., CoT | Gemini 1.5 P., CoT-SC | 0.916 | 0.896 − 0.935 | 52.4 |
| " | Gemini 1.5 P., RAP | 0.847 | 0.822 − 0.871 | 48.7 |
| Gemini 1.5 P., CoT-SC | Gemini 1.5 P., RAP | 0.855 | 0.830 − 0.880 | 48.6 |

Table S.2: A detailed report of the Fleiss Kappa tests [5] on the repeated measurements. It includes 12 approaches, consisting of 3 LLMs (i.e., GPT-4o [2], Claude 3.5 Sonnet [3], and Gemini 1.5 Pro [4]) and 4 prompting techniques (i.e., zero-shot (ZS), chain of thought (CoT), chain of thought with self consistency (CoT-SC), and retrieval augmented prompting (RAP)). All the P-values were approximately zero.

|  |  | Fleiss' Kappa | 95% CI | Std. Error | Z value |
|---|---|---|---|---|---|
| GPT-4o | ZS | 0.980 | $0.973 - 0.987$ | 0.003 | 174 |
|  | CoT | 0.969 | $0.960 - 0.977$ | 0.004 | 172 |
|  | CoT-SC | 0.977 | $0.969 - 0.985$ | 0.004 | 173 |
|  | RAP | 0.985 | $0.979 - 0.991$ | 0.003 | 175 |
| Claude 3.5 S. | ZS | 0.987 | $0.981 - 0.992$ | 0.003 | 175 |
|  | CoT | 0.975 | $0.967 - 0.983$ | 0.004 | 173 |
|  | CoT-SC | 0.982 | $0.975 - 0.988$ | 0.003 | 174 |
|  | RAP | 0.977 | $0.970 - 0.985$ | 0.004 | 174 |
| Gemini 1.5 P. | ZS | 0.996 | $0.993 - 0.999$ | 0.001 | 177 |
|  | CoT | 0.902 | $0.887 - 0.917$ | 0.008 | 165 |
|  | CoT-SC | 0.938 | $0.926 - 0.951$ | 0.006 | 168 |
|  | RAP | 0.991 | $0.987 - 0.996$ | 0.002 | 179 |

Table S.3: Instructions and examples of the prompts used in this study.

| Prompting technique | Instruction |
|---|---|
| Zero Shot prompting | Solve the following multiple choice question with no explanation. Output a single option as the final answer and enclosed by xml tags <answer></answer><br>{Question} {Multiple choices} |
| Chain of Thought prompting * | Solve the following multiple choice question in a step-by-step fashion, starting by summarizing the available information. Output a single option as the final answer and enclosed by xml tags <answer></answer><br>{Question} {Multiple choices} |
| Retrieval Augmented Prompting | Solve the following multiple choice question based on the information provided below. Output a single option as the final answer and enclosed by xml tags <answer></answer><br>{Question} {Multiple choices}<br>{similar chunks} |

* The same instruction was also utilized for the Chain of Thought with Self Consistency prompting technique.

Table S.4: References used for the retrieval augmented prompting (RAP)

| # | Reference | Domain | Organization |
|---|-----------|--------|--------------|
| 1 | Welcome To The Nutrition Care Manual - Diet Manual [6] | Nutrition Care for Individuals and Groups | Academy of Nutrition and Dietetics |
| 2 | Welcome To The Nutrition Care Manual [7] | Nutrition Care for Individuals and Groups | Academy of Nutrition and Dietetics |
| 3 | Standards Of Care In Diabetes - 2024 [8] | Nutrition Care for Individuals and Groups | American Diabetes Association |
| 4 | ASPEN Pocket Cards Guidelines For The Provision And Assessment Of Nutrition Support Therapy In The Pediatric Critically Ill Patient: SCCM And ASPEN [9] | Nutrition Care for Individuals and Groups | American Society for Parenteral and Enteral Nutrition |
| 5 | Guidelines For The Provision Of Nutrition Support Therapy In The Adult Critically Ill Patient [10] | Nutrition Care for Individuals and Groups | American Society for Parenteral and Enteral Nutrition |
| 6 | 2019 ACC/AHA Guideline On The Primary Prevention Of Cardiovascular Disease: A Report Of The American College Of Cardiology/American Heart Association Task Force On Clinical Practice Guidelines [11] | Nutrition Care for Individuals and Groups | American Heart Association |
| 7 | ASPEN Clinical Guidelines: Hyperglycemia And Hypoglycemia In The Neonate Receiving Parenteral Nutrition [12] | Nutrition Care for Individuals and Groups | American Society for Parenteral and Enteral Nutrition |
| 8 | Position Of The Society For Nutrition Education And Behavior: Nutrition Educator Competencies For Promoting Healthy Individuals, Communities, And Food Systems: Rationale And Application [13] | Nutrition Care for Individuals and Groups, Management of Food and Nutrition Programs and Services | United States Department of Agriculture |
| 9 | ASPEN Clinical Guidelines: Nutrition Support Therapy During Adult Anticancer Treatment And In Hematopoietic Cell Transplantation [14] | Nutrition Care for Individuals and Groups | American Society for Parenteral and Enteral Nutrition |

Table S.4: References used for the retrieval augmented prompting (RAP) (continued)

| # | Reference | Domain | Organization |
|---|-----------|--------|--------------|
| 10 | Position Of The Academy Of Nutrition And Dietetics: Benchmarks For Nutrition In Child Care [15] | Principles of Dietetics, Nutrition Care for Individuals and Groups, Management of Food and Nutrition Programs and Services | Academy of Nutrition and Dietetics |
| 11 | Guidelines For Childhood Obesity Prevention Programs: Promoting Healthy Weight In Children [16] | Nutrition Care for Individuals and Groups, Management of Food and Nutrition Programs and Services | United States Department of Agriculture |
| 12 | ASPEN Clinical Guidelines: Parenteral Nutrition Ordering, Order Review, Compounding, Labeling, And Dispensing [17] | Nutrition Care for Individuals and Groups | American Society for Parenteral and Enteral Nutrition |
| 13 | ASPEN Clinical Guidelines: Nutrition Support In Adult Acute And Chronic Renal Failure [18] | Nutrition Care for Individuals and Groups | American Society for Parenteral and Enteral Nutrition |
| 14 | Position Of The Society For Nutrition Education And Behavior: Food And Nutrition Insecurity Among College Students [19] | Nutrition Care for Individuals and Groups, Management of Food and Nutrition Programs and Services | United States Department of Agriculture |
| 15 | Data Table Of BMI-for-age Charts [20] | Nutrition Care for Individuals and Groups | CDC - National Center for Health Statistics |
| 16 | Data Table Of Infant Weight-for-age Charts [21] | Nutrition Care for Individuals and Groups | CDC - National Center for Health Statistics |
| 17 | Data Table Of Infant Head Circumference-for-age Charts [22] | Nutrition Care for Individuals and Groups | CDC - National Center for Health Statistics |
| 18 | Data Table Of Infant Length-for-age Charts [23] | Nutrition Care for Individuals and Groups | CDC - National Center for Health Statistics |
| 19 | Data Table Of Weight-for-age Charts [24] | Nutrition Care for Individuals and Groups | CDC - National Center for Health Statistics |
| 20 | Data Table Of Infant Weight-for-length Charts [25] | Nutrition Care for Individuals and Groups | CDC - National Center for Health Statistics |

Table S.4: References used for the retrieval augmented prompting
(RAP) (continued)

| # | Reference | Domain | Organization |
|---|-----------|--------|--------------|
| 21 | Data Table Of Stature-for-age Charts [26] | Nutrition Care for Individuals and Groups | CDC - National Center for Health Statistics |
| 22 | Data Table Of Weight-for-stature Charts [27] | Nutrition Care for Individuals and Groups | CDC - National Center for Health Statistics |
| 23 | ASPEN Clinical Guidelines: Nutrition Support Of Hospitalized Adult Patients With Obesity [28] | Nutrition Care for Individuals and Groups | American Society for Parenteral and Enteral Nutrition |
| 24 | Guidelines For The Provision Of Nutrition Support Therapy In The Adult Critically Ill Patient: The American Society For Parenteral And Enteral Nutrition [29] | Nutrition Care for Individuals and Groups | American Society for Parenteral and Enteral Nutrition |
| 25 | Position Of The Academy Of Nutrition And Dietetics: Individualized Nutrition Approaches For Older Adults: Long-term Care, Post-acute Care, And Other Settings [30] | Principles of Dietetics, Nutrition Care for Individuals and Groups, Management of Food and Nutrition Programs and Services | Academy of Nutrition and Dietetics |
| 26 | Clinical Guidelines For The Use Of Parenteral And Enteral Nutrition In Adult And Pediatric Patients: Applying The GRADE System To Development Of ASPEN Clinical Guidelines [31] | Nutrition Care for Individuals and Groups | American Society for Parenteral and Enteral Nutrition |
| 27 | Position Of The Academy Of Nutrition And Dietetics: The Role Of Medical Nutrition Therapy And Registered Dietitian Nutritionists In The Prevention And Treatment Of Prediabetes And Type 2 Diabetes [32] | Principles of Dietetics, Nutrition Care for Individuals and Groups, Management of Food and Nutrition Programs and Services | Academy of Nutrition and Dietetics |
| 28 | ASPEN Clinical Guidelines: Nutrition Support Of Neonatal Patients At Risk For Necrotizing Enterocolitis [33] | Nutrition Care for Individuals and Groups | American Society for Parenteral and Enteral Nutrition |
| 29 | Enforce Existing Policies To Soft Drink Products Being Sold And Offered As Promotional Giveaways In Schools And Encourage Additional Governance Regarding Access [34] | Management of Food and Nutrition Programs and Services | United States Department of Agriculture |

Table S.4: References used for the retrieval augmented prompting (RAP) (continued)

| # | Reference | Domain | Organization |
|---|-----------|--------|--------------|
| 30 | Position Of The Academy Of Nutrition And Dietetics: Interprofessional Education In Nutrition As An Essential Component Of Medical Education [35] | Principles of Dietetics, Nutrition Care for Individuals and Groups, Management of Food and Nutrition Programs and Services | Academy of Nutrition and Dietetics |
| 31 | Position Of The Academy Of Nutrition And Dietetics, Society For Nutrition Education And Behavior, And School Nutrition Association: Comprehensive Nutrition Programs And Services In Schools [36] | Principles of Dietetics, Nutrition Care for Individuals and Groups, Management of Food and Nutrition Programs and Services | Academy of Nutrition and Dietetics |
| 32 | Prevention Of Pediatric Overweight And Obesity: Position Of The Academy Of Nutrition And Dietetics Based On An Umbrella Review Of Systematic Reviews [37] | Principles of Dietetics, Nutrition Care for Individuals and Groups, Management of Food and Nutrition Programs and Services | Academy of Nutrition and Dietetics |
| 33 | Position Of The Academy Of Nutrition And Dietetics: Food Insecurity In The United States [38] | Principles of Dietetics, Nutrition Care for Individuals and Groups, Management of Food and Nutrition Programs and Services | Academy of Nutrition and Dietetics |
| 34 | Dietary Reference Intakes: Applications In Dietary Assessment [39] | Nutrition Care for Individuals and Groups, Foodservice Systems | Institue of Medicine |
| 35 | Dietary Reference Intakes For Calcium And Vitamin D [40] | Nutrition Care for Individuals and Groups, Foodservice Systems | Institue of Medicine |
| 36 | Dietary Reference Intakes For Calcium, Phosphorus, Magnesium, Vitamin D, And Fluoride [41] | Nutrition Care for Individuals and Groups, Foodservice Systems | Institue of Medicine |
| 37 | Dietary Reference Intakes For Energy, Carbohydrate, Fiber, Fat, Fatty Acids, Cholesterol, Protein, And Amino Acids [42] | Nutrition Care for Individuals and Groups, Foodservice Systems | Institue of Medicine |

Continued on next page

Table S.4: References used for the retrieval augmented prompting (RAP) (continued)

| # | Reference | Domain | Organization |
|---|-----------|--------|--------------|
| 38 | Dietary Reference Intakes For Thiamin, Riboflavin, Niacin, Vitamin B6, Folate, Vitamin B12, Pantothenic Acid, Biotin, And Choline [43] | Nutrition Care for Individuals and Groups, Foodservice Systems | Institue of Medicine |
| 39 | Dietary Reference Intakes For Vitamin A, Vitamin K, Arsenic, Boron, Chromium, Copper, Iodine, Iron, Manganese, Molybdenum, Nickel, Silicon, Vanadium, And Zinc [44] | Nutrition Care for Individuals and Groups, Foodservice Systems | Institue of Medicine |
| 40 | Dietary Reference Intakes For Vitamin C, Vitamin E, Selenium, And Carotenoids [45] | Nutrition Care for Individuals and Groups, Foodservice Systems | Institue of Medicine |
| 41 | Dietary Reference Intakes For Water, Potassium, Sodium, Chloride, And Sulfate [46] | Nutrition Care for Individuals and Groups, Foodservice Systems | Institue of Medicine |
| 42 | ASPEN Clinical Guidelines: Nutrition Support Of Neonates Supported With Extracorporeal Membrane Oxygenation [47] | Nutrition Care for Individuals and Groups | American Society for Parenteral and Enteral Nutrition |
| 43 | Soft Drink Resolution [48] | Nutrition Care for Individuals and Groups, Management of Food and Nutrition Programs and Services | United States Department of Agriculture |
| 44 | ASPEN Clinical Guidelines: Nutrition Support Of Hospitalized Pediatric Patients With Obesity [49] | Nutrition Care for Individuals and Groups | American Society for Parenteral and Enteral Nutrition |
| 45 | Treatment Of Pediatric Overweight And Obesity: Position Of The Academy Of Nutrition And Dietetics Based On An Umbrella Review Of Systematic Reviews [50] | Principles of Dietetics, Nutrition Care for Individuals and Groups, Management of Food and Nutrition Programs and Services | Academy of Nutrition and Dietetics |
| 46 | 2021 Guideline For The Prevention Of Stroke In Patients With Stroke And Transient Ischemic Attack: A Guideline From The American Heart Association /slash American Stroke Association [51] | Nutrition Care for Individuals and Groups | American Heart Association |

Continued on next page

Table S.4: References used for the retrieval augmented prompting (RAP) (continued)

| # | Reference | Domain | Organization |
|---|-----------|--------|--------------|
| 47 | American Society For Parenteral And Enteral Nutrition Guidelines For The Selection And Care Of Central Venous Access Devices For Adult Home Parenteral Nutrition Administration [52] | Nutrition Care for Individuals and Groups | American Society for Parenteral and Enteral Nutrition |
| 48 | ASPEN-FELANPE Clinical Guidelines: Nutrition Support Of Adult Patients With Enterocutaneous Fistula [53] | Nutrition Care for Individuals and Groups | American Society for Parenteral and Enteral Nutrition |
| 49 | Resolution To Support Nutrition Labeling And Nutritionally Improved Menu Offerings In Fast-Food And Other Chain Restaurants [54] | Management of Food and Nutrition Programs and Services | United States Department of Agriculture |
| 50 | Resolution To Support Responsible Food Marketing To Children [55] | Management of Food and Nutrition Programs and Services | United States Department of Agriculture |
| 51 | Position Of The Academy Of Nutrition And Dietetics: Micronutrient Supplementation [56] | Principles of Dietetics, Nutrition Care for Individuals and Groups, Management of Food and Nutrition Programs and Services | Academy of Nutrition and Dietetics |
| 52 | Guidelines For The Provision And Assessment Of Nutrition Support Therapy In The Adult Critically Ill Patient [57] | Nutrition Care for Individuals and Groups | American Society for Parenteral and Enteral Nutrition |
| 53 | ASPEN Clinical Guidelines: Nutrition Support Of Adult Patients With Hyperglycemia [58] | Nutrition Care for Individuals and Groups | American Society for Parenteral and Enteral Nutrition |
| 54 | ASPEN Clinical Guidelines: Nutrition Support Of The Critically Ill Child [59] | Nutrition Care for Individuals and Groups | American Society for Parenteral and Enteral Nutrition |
| 55 | Guidelines For The Provision And Assessment Of Nutrition Support Therapy In The Pediatric Critically Ill Patient: Society Of Critical Care Medicine And American Society For Parenteral And Enteral Nutrition [60] | Nutrition Care for Individuals and Groups | American Society for Parenteral and Enteral Nutrition |

Table S.4: References used for the retrieval augmented prompting (RAP) (continued)

| # | Reference | Domain | Organization |
|---|-----------|--------|--------------|
| 56 | Safe Practices For Parenteral Nutrition [61] | Nutrition Care for Individuals and Groups | American Society for Parenteral and Enteral Nutrition |
| 57 | ASPEN Clinical Guidelines: Nutrition Screening, Assessment, And Intervention In Adults [62] | Nutrition Care for Individuals and Groups | American Society for Parenteral and Enteral Nutrition |
| 58 | Dietary Reference Intakes For Sodium And Potassium [63] | Nutrition Care for Individuals and Groups, Foodservice Systems | Institue of Medicine |
| 59 | ASPEN Clinical Guidelines: Nutrition Support Of Neonatal Patients At Risk For Metabolic Bone Disease [64] | Nutrition Care for Individuals and Groups | American Society for Parenteral and Enteral Nutrition |
| 60 | Guidelines For Parenteral Nutrition In Preterm Infants: The American Society For Parenteral And Enteral Nutrition [65] | Nutrition Care for Individuals and Groups | American Society for Parenteral and Enteral Nutrition |
| 61 | American Cancer Society Guideline For Diet And Physical Activity For Cancer Prevention [66] | Nutrition Care for Individuals and Groups | American Cancer Society |
| 62 | Position Of The Society For Nutrition Education And Behavior: The Importance Of Including Environmental Sustainability In Dietary Guidance [67] | Nutrition Care for Individuals and Groups, Management of Food and Nutrition Programs and Services | United States Department of Agriculture |
| 63 | Position Of The Society For Nutrition Education And Behavior: Healthful Food For Children Is The Same As Adults [68] | Nutrition Care for Individuals and Groups, Management of Food and Nutrition Programs and Services | United States Department of Agriculture |
| 64 | Position Of The Academy Of Nutrition And Dietetics: Child And Adolescent Federally Funded Nutrition Assistance Programs [69] | Principles of Dietetics, Nutrition Care for Individuals and Groups, Management of Food and Nutrition Programs and Services | Academy of Nutrition and Dietetics |

Table S.4: References used for the retrieval augmented prompting (RAP) (continued)

| # | Reference | Domain | Organization |
|---|-----------|--------|--------------|
| 65 | Position Of The Academy Of Nutrition And Dietetics: Nutrition Informatics [70] | Principles of Dietetics, Nutrition Care for Individuals and Groups, Management of Food and Nutrition Programs and Services | Academy of Nutrition and Dietetics |
| 66 | ASPEN Clinical Guidelines: Nutrition Support Of Children With Human Immunodeficiency Virus Infection [71] | Nutrition Care for Individuals and Groups | American Society for Parenteral and Enteral Nutrition |
| 67 | Position Of The Academy Of Nutrition And Dietetics And The Society For Nutrition Education And Behavior: Food And Nutrition Programs For Community-residing Older Adults [72] | Nutrition Care for Individuals and Groups, Management of Food and Nutrition Programs and Services | United States Department of Agriculture |
| 68 | American Society For Parenteral And Enteral Nutrition Clinical Guidelines: The Validity Of Body Composition Assessment In Clinical Populations [73] | Nutrition Care for Individuals and Groups | American Society for Parenteral and Enteral Nutrition |
| 69 | Position Of The Academy Of Nutrition And Dietetics: Malnutrition (undernutrition) Screening Tools For All Adults [74] | Principles of Dietetics, Nutrition Care for Individuals and Groups, Management of Food and Nutrition Programs and Services | Academy of Nutrition and Dietetics |
| 70 | List Of Resolutions Passed By The Membership: 1970-1992 [75] | Management of Food and Nutrition Programs and Services | United States Department of Agriculture |
| 71 | SNEB Healthy Meeting Guidelines [76] | Management of Food and Nutrition Programs and Services | United States Department of Agriculture |
| 72 | Building A Nutrition Education Evidence Data Base To Support Policy And Planning In Developing Countries [77] | Management of Food and Nutrition Programs and Services | United States Department of Agriculture |
| 73 | Recommit To An Ongoing Lifespan Approach And Address The Needs Of A Growing Aging Population [78] | Management of Food and Nutrition Programs and Services | United States Department of Agriculture |

Table S.4: References used for the retrieval augmented prompting (RAP) (continued)

| # | Reference | Domain | Organization |
|---|-----------|--------|--------------|
| 74 | Diversity, Equity, And Inclusion Statement For The Society Of Nutrition Education And Behavior [79] | Management of Food and Nutrition Programs and Services | United States Department of Agriculture |
| 75 | Position Of The Academy Of Nutrition And Dietetics And The Society For Nutrition Education And Behavior: Food And Nutrition Programs For Community-Residing Older Adults [80] | Principles of Dietetics, Nutrition Care for Individuals and Groups, Management of Food and Nutrition Programs and Services | Academy of Nutrition and Dietetics |
| 76 | State Of Nutrition Education & Promotion For Children & Adolescents [81] | Nutrition Care for Individuals and Groups, Management of Food and Nutrition Programs and Services | United States Department of Agriculture |
| 77 | How To Comply With The Americans With Disabilities Act: A Guide For Restaurants And Other Food Service Employers [82] | Management of Food and Nutrition Programs and Services | U.S. Equal Employment Opportunity Commission |
| 78 | HACCP Principles & Application Guidelines [83] | Foodservice Systems | U.S. Food and Drug Adimindistration |
| 79 | Generally Recognized As Safe (GRAS) [84] | Foodservice Systems | U.S. Food and Drug Adimindistration |
| 80 | Small Entity Compliance Guide On Structure/Function Claims [85] | Principles of Dietetics | U.S. Food and Drug Admistration |
| 81 | Food Code [86] | Foodservice Systems | U.S. Food and Drug Adimindistration |
| 82 | Organic 101: What The USDA Organic Label Means [87] | Principles of Dietetics | United States Department of Agriculture |
| 83 | National School Lunch Program Factsheet [88] | Management of Food and Nutrition Programs and Services | United States Department of Agriculture |
| 84 | School Breakfast Program Factsheet [89] | Management of Food and Nutrition Programs and Services | United States Department of Agriculture |

Table S.4: References used for the retrieval augmented prompting (RAP) (continued)

| # | Reference | Domain | Organization |
|---|-----------|--------|--------------|
| 85 | Healthy Eating For Preschoolers [90] | Nutrition Care for Individuals and Groups | United States Department of Agriculture - Food and Nutrition Service |
| 86 | Start Simple With MyPlate [91] | Nutrition Care for Individuals and Groups | United States Department of Agriculture - Food and Nutrition Service |
| 87 | Understanding Food Quality Labels: A Guide To AMS Grade Shields, Value-Added Labels, And Official Seals [92] | Principles of Dietetics | United States Department of Agriculture |
| 88 | Dietary Guidelines For Americans (2020 - 2025) [93] | Nutrition Care for Individuals and Groups | Dietary Guidelines for Americans |
| 89 | Make Breakfast First Class [94] | Management of Food and Nutrition Programs and Services | United States Department of Agriculture |
| 90 | Be Salt Smart [95] | Nutrition Care for Individuals and Groups | United States Department of Agriculture - Food and Nutrition Service |
| 91 | Make Better Beverage Choices [96] | Nutrition Care for Individuals and Groups | United States Department of Agriculture - Food and Nutrition Service |
| 92 | Healthy Snacking With MyPlate [97] | Nutrition Care for Individuals and Groups | United States Department of Agriculture - Food and Nutrition Service |
| 93 | Healthy Eating For Women Who Are Pregnant Or Breastfeeding [98] | Nutrition Care for Individuals and Groups | United States Department of Agriculture - Food and Nutrition Service |
| 94 | Healthy Eating For Infants [99] | Nutrition Care for Individuals and Groups | United States Department of Agriculture - Food and Nutrition Service |

Table S.4: References used for the retrieval augmented prompting (RAP) (continued)

| # | Reference | Domain | Organization |
|---|-----------|--------|--------------|
| 95 | Healthy Eating For Toddlers [100] | Nutrition Care for Individuals and Groups | United States Department of Agriculture - Food and Nutrition Service |
| 96 | Healthy Eating For Kids [101] | Nutrition Care for Individuals and Groups | United States Department of Agriculture - Food and Nutrition Service |
| 97 | Healthy Eating For Teens [102] | Nutrition Care for Individuals and Groups | United States Department of Agriculture - Food and Nutrition Service |
| 98 | Focus On Whole Fruits [103] | Nutrition Care for Individuals and Groups | United States Department of Agriculture - Food and Nutrition Service |
| 99 | Healthy Eating For Adults [104] | Nutrition Care for Individuals and Groups | United States Department of Agriculture - Food and Nutrition Service |
| 100 | Healthy Eating For Older Adults [105] | Nutrition Care for Individuals and Groups | United States Department of Agriculture - Food and Nutrition Service |
| 101 | Healthy Eating For Families [106] | Nutrition Care for Individuals and Groups | United States Department of Agriculture - Food and Nutrition Service |
| 102 | Healthy Eating For Young Adults [107] | Nutrition Care for Individuals and Groups | United States Department of Agriculture - Food and Nutrition Service |
| 103 | Meal Planning [108] | Nutrition Care for Individuals and Groups | United States Department of Agriculture - Food and Nutrition Service |

Table S.4: References used for the retrieval augmented prompting (RAP) (continued)

| # | Reference | Domain | Organization |
|---|-----------|--------|--------------|
| 104 | Cut Back On Added Sugars [109] | Nutrition Care for Individuals and Groups | United States Department of Agriculture - Food and Nutrition Service |
| 105 | Rethink Fats [110] | Nutrition Care for Individuals and Groups | United States Department of Agriculture - Food and Nutrition Service |
| 106 | Enjoy Vegetarian Meals [111] | Nutrition Care for Individuals and Groups | United States Department of Agriculture - Food and Nutrition Service |
| 107 | Move To Low-Fat Or Fat-Free Dairy [112] | Nutrition Care for Individuals and Groups | United States Department of Agriculture - Food and Nutrition Service |
| 108 | Eat Healthy On A Budget [113] | Nutrition Care for Individuals and Groups | United States Department of Agriculture - Food and Nutrition Service |
| 109 | Make Half Your Grains Whole Grains [114] | Nutrition Care for Individuals and Groups | United States Department of Agriculture - Food and Nutrition Service |
| 110 | Vary Your Vegetables [115] | Nutrition Care for Individuals and Groups | United States Department of Agriculture - Food and Nutrition Service |
| 111 | Vary Your Protein Routine [116] | Nutrition Care for Individuals and Groups | United States Department of Agriculture - Food and Nutrition Service |
| 112 | Celebrations And Gatherings [117] | Nutrition Care for Individuals and Groups | United States Department of Agriculture - Food and Nutrition Service |

Table S.4: References used for the retrieval augmented prompting (RAP) (continued)

| # | Reference | Domain | Organization |
|---|-----------|--------|--------------|
| 113 | Dine Out /Take Out [118] | Nutrition Care for Individuals and Groups | United States Department of Agriculture - Food and Nutrition Service |
| 114 | Healthy Food Prep [119] | Nutrition Care for Individuals and Groups | United States Department of Agriculture - Food and Nutrition Service |
| 115 | Kitchen Time-Savers [120] | Nutrition Care for Individuals and Groups | United States Department of Agriculture - Food and Nutrition Service |
| 116 | Grocery Shopping [121] | Nutrition Care for Individuals and Groups | United States Department of Agriculture - Food and Nutrition Service |
| 117 | ASPEN Clinical Guidelines: Support Of Pediatric Patients With Intestinal Failure At Risk Of Parenteral Nutrition–Associated Liver Disease [122] | Nutrition Care for Individuals and Groups | American Society for Parenteral and Enteral Nutrition |
| 118 | WHO Child Growth Standards: Head Circumference-for-age, Arm Circumference-for-age, Triceps Skinfold-for-age And Subscapular Skinfold-for-age: Methods And Development [123] | Nutrition Care for Individuals and Groups | World Health Organization |
| 119 | WHO Child Growth Standards: Length/height-for-age, Weight-for-age, Weight-for-length, Weight-for-height And Body Mass Index-for-age: Methods And Development [124] | Nutrition Care for Individuals and Groups | World Health Organization |
| 120 | WHO Child Growth Standards: Growth Velocity Based On Weight, Length And Head Circumference: Methods And Development [125] | Nutrition Care for Individuals and Groups | World Health Organization |

Table S.4: References used for the retrieval augmented prompting (RAP) (continued)

| # | Reference | Domain | Organization |
|---|-----------|--------|--------------|
| 121 | 2023 AHA/ACC/ACCP/ASPC/NLA/PCNA Guideline For The Management Of Patients With Chronic Coronary Disease: A Report Of The American Heart Association/American College Of Cardiology Joint Committee On Clinical Practice Guidelines [126] | Nutrition Care for Individuals and Groups | American Heart Association |
| 122 | Position Of The American Dietetic Association, Society For Nutrition Education, And American School Food Service Association: Nutrition Services: An Essential Component Of Comprehensive School Health Programs [127] | Nutrition Care for Individuals and Groups, Management of Food and Nutrition Programs and Services | United States Department of Agriculture |
| 123 | Clinical Guidelines For The Use Of Parenteral And Enteral Nutrition In Adult And Pediatric Patients, 2009 [128] | Nutrition Care for Individuals and Groups | American Society for Parenteral and Enteral Nutrition |
| 124 | Practice Is Evolving-a Briefing Of The Revised 2024 Scopes And Standards Of Practice [129] | Management of Food and Nutrition Programs and Services | Academy of Nutrition and Dietetics |
| 125 | Correction [130] | Nutrition Care for Individuals and Groups | American Society for Parenteral and Enteral Nutrition |



\end{document}